\newcommand{\arXiv}{}
\ifdefined\arXiv
\documentclass[11pt]{article}

\pdfoutput=1

\usepackage[margin=1in]{geometry}

\usepackage{authblk}
\usepackage{subcaption}

\else
\documentclass[sigconf]{acmart}\usepackage{etoolbox}

\AtBeginDocument{%
  }

\setcopyright{acmlicensed}
\copyrightyear{2024}
\acmYear{2024}
\acmDOI{XXXXXXX.XXXXXXX}
\fi

\usepackage{epsfig}
\usepackage{amsfonts}
\usepackage{balance}  

\usepackage[numbers]{natbib}
\usepackage{tabularx}
\usepackage{amsmath}

\usepackage{booktabs} 
\usepackage{mathtools} 

\usepackage{changepage}
\usepackage{algpseudocode}
\usepackage{bm}
\usepackage{graphicx}
\usepackage{comment}
\usepackage{array}
\ifdefined\VLDB
\newcommand{\subparagraph}{}
\fi
\usepackage{amsthm}
\usepackage{algorithm}
\usepackage{blindtext}
\usepackage{multirow}
\usepackage{yfonts}
\usepackage{xcolor}
\usepackage{url}
\usepackage{balance}
\usepackage{nopageno}
\usepackage{enumitem}
\usepackage{dsfont}
\usepackage{thmtools}
\usepackage{thm-restate}

\usepackage[tableposition=top]{caption}
\usepackage[labelfont=bf]{caption}

\usepackage{tikz}

\usepackage{hyperref}

\input{sections/figures/tikz-macros_new}

\newtheorem*{theorem*}{Theorem}

\algtext*{EndFunction}
\algtext*{EndIf}
\algtext*{EndFor}
\algtext*{EndWhile}
\algtext*{EndIf}

\newif\ifcomm
\commtrue
\ifcomm
\else
\commfalse
\fi
\ifcomm
	\newcommand{\mycomm}[3]{{\footnotesize{{\color{#2} \textbf{[#1: #3]}}}}}
	\newcommand{\CRdel}[1]{\textcolor{red}{\sout{#1}}}
\else
    \newcommand{\mycomm}[3]{}
    \newcommand{\CRdel}[1]{}
\fi
\newcommand{\rana}[1]{\mycomm{Rana}{blue}{#1}}

\begin{document}

\floatstyle{plaintop}
\restylefloat{table}


\title{Queueing, Predictions, and LLMs: Challenges and Open Problems}

\author[1]{Michael Mitzenmacher}

\author[1]{Rana Shahout}

\affil[1]{Harvard University, USA}

\date{}

\ifdefined\arXiv
\maketitle
\fi

\begin{abstract}

Queueing systems present many opportunities for applying machine-learning predictions, such as estimated service times, to improve system performance. This integration raises numerous open questions about how predictions can be effectively leveraged to improve scheduling decisions.
Recent studies explore queues with predicted service times, typically aiming to minimize job time in the system. We review these works, highlight the effectiveness of predictions, and present open questions on queue performance.

We then move to consider an important practical example of using predictions in scheduling, namely Large Language Model (LLM) systems, which presents novel scheduling challenges and highlights the potential for predictions to improve performance. In particular, we consider LLMs performing inference. 
Inference requests (jobs) in LLM systems are inherently complex; they have variable inference times, dynamic memory footprints that are constrained by key-value (KV) store memory limitations, and multiple possible preemption approaches that affect performance differently. We provide background on the important aspects of scheduling in LLM systems, and introduce new models and open problems that arise from them. We argue that there are significant opportunities for applying insights and analysis from queueing theory to scheduling in LLM systems.

\end{abstract}

\maketitle
\section{Introduction}

In this paper, we 
survey recent work
on using predictions in queueing systems,
as well as recent work on the specific setting of scheduling in Large Language Model (LLM) systems, where predictions seem both useful and natural.
We focus on presenting several open questions for consideration.  Our purpose is to highlight the work in these areas, and encourage researchers to tackle the many interesting problems raised by systems that make use of predictions.\footnote{We note that this survey is written in conjunction with a tutorial the authors will give at SIGMETRICS 2025, and acknowledge that the survey includes discussion of work conducted by the authors, alongside other contributions.}

To introduce the queueing theoretic problems, let us consider a standard queue,
such as an M/G/1 queue -- where jobs arrive to a single-server queue, according to a Poisson arrival process with general i.i.d service times. There are two common scheduling approaches, depending on the information available.  First, it may be that no information about any job's service time is known, in which case First-In First-Out (FIFO), also referred to as First-Come First-Served (FCFS), is generally used. Second, it may be that every job's service time is known, in which case the Shortest Remaining Processing Time (SRPT) optimizes the expected response time. 

In practice, however, there are many settings where the exact service time is not known in advance. A potentially promising approach for such settings is to utilize predictions, which may be generated by machine learning models; these models can estimate service times and inform scheduling decisions.
Several recent studies have explored queues that use predicted, rather than exact, service times to reduce the average response time \cite{akbari2023seh,dell2015psbs,dell2014revisiting,mailach2017scheduling,mitzenmacher2021queues,ScullyGM22,wierman2008scheduling}. Recent work has also considered the setting of scheduling jobs with deadlines \cite{salman2023scheduling,salman2023evaluating}, demonstrating that predictions can improve scheduling decisions in that setting as well.

 More generally, the integration of predictions with algorithm design has given rise to the field of ``algorithms with predictions,'' also known as learning-augmented algorithms. In this area, classical algorithms are improved by incorporating advice or predictions from machine learning models (or other sources), ideally with provable performance guarantees. 
 While the idea of using additional information to improve algorithms and data structures has some history -- the study of online algorithms with advice is a notable example \cite{boyar2017online} -- the specific motivation to make use of predictions from machine learning algorithms has led to new definitions, models, and results.
 Learning-augmented algorithms have demonstrated their effectiveness across a range of areas, as shown in the collection of papers \cite{githubio} on the subject and as discussed in the surveys \cite{DBLP:books/cu/20/MitzenmacherV20,DBLP:journals/cacm/MitzenmacherV22}.  

There are many important and basic queueing theory questions that arise once one focuses on predicted service time. Accordingly, our first goal in this paper is to survey some of the recent work on using predictions for scheduling in single queues and queueing systems, and describe open problems and research directions in this area. 

From this starting point, we then consider a more concrete and timely application setting: LLM systems.
LLM systems have transformed many domains by enabling powerful AI-driven applications, rapidly integrating into workflows and decision-making processes. 
These systems consist of two main phases: training and inference. Training is a computationally intensive offline process where models learn from massive datasets, while inference, the focus of this work, is the real-time execution phase where pre-trained models generate responses to user requests.
At inference, LLM models generate text token (usually a word or a part of a word) by token, each new token relying on the context of previously generated tokens in an autoregressive process (where each output is fed back as input for the next prediction).
Given the increasing availability of open-source models such as Llama~\cite{touvron2023llama} and DeepSeek~\cite{guo2025deepseek,liu2024deepseek}, inference has become the most common mode of interaction with LLMs, and that will be our focus in this work.  

LLM systems present a wealth of new queueing and scheduling challenges that expand on problems from traditional queueing systems. 
While modeling LLM systems remains largely unexplored in the queueing community, we believe queueing theory insights may lead to better scheduling systems.
We briefly describe some of the problems and issues in LLM scheduling here, and elaborate on them later in the paper. 

One issue for LLM systems is that they operate with multiple goals: cost (e.g., computational and financial) and response quality play a critical role alongside traditional performance metrics like latency and throughput. Optimizing across goals leads to interesting challenges.

Another challenge LLM inference introduces not present in standard queueing models is the need for an ever-growing key-value (KV) cache~\cite{pope2023efficiently}.
The KV cache is crucial for reusing intermediate computations, improving efficiency by avoiding redundant recalculations at each step of autoregressive token generation. 
In autoregressive models, each token is generated sequentially, and each new token is conditioned on all previously generated tokens. Thus, the cache accumulates more data as the response grows. However, its presence introduces memory constraints that complicate scheduling, as each request consumes GPU memory that cannot be freed until the request is completed. 

Moreover, preemption, where a running job can be interrupted to prioritize a more urgent request, is non-trivial in LLM inference due to KV cache management. In LLM systems, preempting a request requires dealing with its allocated KV cache memory, which either must be kept in GPU memory (which uses space other jobs may need), deleted from memory (which requires recomputation),  or transferred to the CPU memory (which incurs some cost in time). The high memory footprint of LLMs, combined with the inefficiencies of frequent cache transfers, makes preemption costly yet necessary to prevent long-running requests from blocking shorter ones and to mitigate head-of-line blocking effects. These complexities highlight the need for scheduling strategies designed explicitly for LLM inference, accounting for memory constraints.

LLM systems also vary in complexity depending on their components and deployment settings.
As an example, requests in LLM inference progress through two distinct phases, a prefill phase and a decode phase.
The prefill phase is compute-bound; the decode phase is memory-bandwidth-bound.  
The scheduler must efficiently balance these phases.
More complex systems like compound AI systems integrate multiple interacting components rather than relying on a single model. LLMs in these systems often issue external API calls for retrieval or computation, introducing new scheduling constraints and the need to decide how to manage the KV cache during API calls. 
Multiple LLM systems provide further problems.
In some settings,
multiple LLMs of different sizes are available, requiring a routing strategy to balance cost, response time, and answer quality. Other systems involve multi-step pipelines, where requests passing through multiple LLMs in sequence create dependencies that influence scheduling.


Finally, new LLM reasoning systems introduce additional scheduling challenges. These systems extend traditional inference by generating structured reasoning steps before reaching a final answer. Reasoning-based systems can benefit from evaluating early results to determine whether continued computation is necessary or if resources should be reallocated to other tasks. 
Additionally, some requests may resolve quickly, while others require extensive exploration. As a result, the notion of request size extends beyond the token count to include the number of reasoning steps required for convergence.

The following sections provide a survey and suggest open problems in these areas.
Section~\ref{sec:alg_w_predictions} surveys recent works on using predictions for scheduling in single queues and queueing systems and outlines open problems and research directions. Section~\ref{sec:llm_background} provides background on LLM systems, discusses their challenges, and explains how they differ from standard queueing models.
Sections~\ref{sec:llm_scheduling}, \ref{sec:compound_systems}, and \ref{sec:llm_reasoning} deal with LLM systems. We examine three types of LLM systems that offer challenges for work in scheduling and load balancing: (1) a single instance of an LLM, (2) compound AI systems, and (3) reasoning LLM systems.  Section~\ref{sec:conclusion} concludes with our summary.
\section{Scheduling with Job Size Predictions}
\label{sec:alg_w_predictions}

\subsection{Extensions of Standard Queueing Models}

A natural starting point for considering scheduling with job size predictions is the standard M/G/1 queue. Typically, such queues use FIFO scheduling when job sizes\footnote{We will use terms like job size, size, service time, and time interchangeably, where the meaning is clear.} are not known, and use SRPT when job sizes are known. These scheduling policies are well understood, as are related policies that use known job sizes, such as shortest job first (SJF) or preemptive shortest job first (PSJF).  It seems natural to extend these standard size-based scheduling policies to the setting where job sizes are not known exactly but are instead predicted.

Mitzenmacher \cite{mitzenmacher2019scheduling} considers a simple model that is in the spirit of traditional queueing theory problems where jobs may have priority classes. Instead of just each job's size being modeled as independently selected from a fixed distribution, the model is extended, so now each job independently has both a size and a predicted size selected from a fixed (two-dimensional) distribution.  That is, there is a density function $g(x,y)$ for the probability that a job has service time $x$ and predicted service time $y$.  
Given this model for predictions, the equations for the expected response time of the variants of SRPT, SJF, and PSJF that use the predicted sizes to schedule jobs are derived. (There is no settled naming convention for these variants, but following \cite{mitzenmacher2019scheduling}, we will refer to these as shortest predicted remaining processing time (SPRPT), shortest predicted job first (SPJF), and preemptive shortest predicted job first (PSPJF).)  

It is not clear what are realistic models for predicted job times. 
The work \cite{mitzenmacher2019scheduling} introduces and studies some artificial prediction distributions that are mathematically natural, including where a job with true job size $x$ has a predicted job size that is exponentially distributed with mean $x$, or uniformly distributed in the range $[(1-\alpha)x,(1+\alpha)x]$ for some parameter $\alpha$.

As an example, Table~\ref{tab:simple_results}, taken from results in \cite{mitzenmacher2019scheduling}, shows the expected response time in equilibrium for SPJF and SPRPT and compares them to the expected response times for FIFO, SJF, and SRPT.  
The primary takeaway from this example is that while using predictions is naturally not as good as using exact information, it provides significant gains over using FIFO, which does not use any information about job times. 

\begin{table}[h]
    \centering
    \begin{tabular}{|c|c|c|c|c|c|c|c|}
        \hline
        $\lambda$ & FIFO & SJF & SPJF & PSJF & PSPJF & SRPT  & SPRPT \\
        \hline
        0.5 & 2.0000 & 1.7127 & 1.7948 & 1.5314 & 1.6636 & 1.4254 & 1.6531 \\
        0.6 & 2.5000 & 1.9625 & 2.1086 & 1.7526 & 1.9527 & 1.6041 & 1.9305 \\
        0.7 & 3.3333 & 2.3122 & 2.5726 & 2.0839 & 2.3970 & 1.8746 & 2.3539 \\
        0.8 & 5.0000 & 2.8822 & 3.3758 & 2.6589 & 3.1943 & 2.3528 & 3.1168 \\
        0.9 & 10.0000 & 4.1969 & 5.3610 & 4.0518 & 5.2232 & 3.5521 & 5.0481 \\
        0.95 & 20.0000 & 6.2640 & 8.6537 & 6.2648 & 8.6166 & 5.5410 & 8.3221 \\
        0.98 & 50.0000 & 11.2849 & 16.9502 & 11.5513 & 17.1090 & 10.4947 & 16.6239 \\
        0.99 & 100.0000 & 18.4507 & 29.0536 & 18.9556 & 29.3783 & 17.6269 & 28.7302 \\
        \hline
    \end{tabular}
    \caption{Results from equations (to four decimal places) for FIFO, Shortest Job First (SJF), Shortest Predicted Job First (SPJF),
    Preemptive Shortest Job First (PSJF), Preemptive Shortest Predicted Job First (PSPJF), Shortest Remaining Processing Time (SRPT), and Shortest Predicted Remaining Processing Time (SPRPT), where the predicted time for a job with size $x$ is exponentially distributed with mean $x$. $\lambda$ is the arrival rate. Taken from \cite{mitzenmacher2019scheduling}.}
    \label{tab:simple_results}
\end{table}

While \cite{mitzenmacher2019scheduling} derives equations for these prediction-based policies directly, following similar previous derivations for the standard variants without predictions (see, e.g. \cite{harchol2013performance}), it should be noted that these particular prediction-based scheduling schemes are amenable to analysis using the SOAP methodology of \cite{scully2018soap}.  (See also \cite{scully2018soapbubbles}.) The SOAP methodology requires that jobs be serviced according to some ranking function that depends only on a job's ``type'' and the amount of time it has been served.  A job's type could correspond to a job class in a system with job classes, or the job's size, or both. In this setting, a job's type corresponds to the pair $(x,y)$ representing its actual and predicted service size. When scheduling by shortest predicted remaining processing time, for example, if a job has been served for $a$ units of time, its rank is simply $y-a$ (the predicted remaining service time) as long as $a \leq x$ (since after being served for time $x$ the job completes). The SOAP methodology provides a very general (albeit sometimes difficult) approach for analyzing M/G/1 queueing variants using predictions, as long as the conditions for using SOAP are satisfied, which is a significant limitation. For example, the SOAP methodology cannot be applied when the scheduling policy depends on how many jobs are in the queue awaiting service.  

\medskip 

\noindent{{\bf Open Questions:}}
\begin{itemize}
\item Are there natural models of predictions and the resulting prediction errors in queueing that are realistic across a range of problems, and/or particularly worthy of future study?
\item Can we design a tool that readily numerically computes results for standard queueing policies for M/G/1 queues given a prediction model in some standardized form?
\item Are there analysis approaches (extending SOAP or otherwise) to deal with more general settings with predictions, such as using predictions only when the number of items in the queue is sufficiently high?
\item The Gittins policy \cite{scully2021gittins} should be optimal in this setting.  Are there conditions under which implementing a Gittins policy would be natural?
\item Here we have described predictions as being real-valued.
A prediction, however, could take the form of a predicted distribution for a job, as opposed to a value (see, e.g., \cite{dinitz2024binary}). The analysis of effective scheduling approaches when predictions take the form of distributions remains open.  
\item Predictors can, for various reasons, possibly become poor or degrade.  Can we model systems where predictions are monitored and possibly ignored if they appear sufficiently incorrect?  In such a system, there may be a mix of predicted and unknown service times.  
\end{itemize}

\subsection{1-Bit Predictions}

Additional recent work continues to develop the use of predictions.  Mitzenmacher \cite{mitzenmacher2021queues} studies the viability of using 1-bit predictions, where the bit corresponds to
prediction leads to a simple implementation: short jobs can be placed at the front of the queue, while long jobs can be placed at the back of the queue. Such an implementation could be preemptive, so a new short job preempts any running job, or not.  

 From the theoretical standpoint, 1-bit predictions lead to a mathematically more tractable system.  For example, consider the standard M/M/1 model with Poisson arrivals of rate $\lambda$ and where jobs have exponentially distributed service times with mean 1, but now we add the exponential prediction model.
 A job with true service time $x$ can be thought of as having a predicted time $y$, where $y$ is exponentially distributed with mean $x$, and if $y > T$ the job is marked as long and otherwise it is marked as short. The response time without preemption is shown to have the form: 
 $$t_1 = \frac{\lambda (1 - \lambda (1 - 2\sqrt{T}K_1(2\sqrt{T})))}
{(1 - \lambda)(1 - \lambda(1 - 2T K_2(2\sqrt{T})))} + 1,$$
and the response time with preemption is shown to be 
$$t_2 = \frac{1 - \lambda + \lambda 2\sqrt{T}K_1(2\sqrt{T})}
{(1 - \lambda)(1 - \lambda(1 - 2T K_2(2\sqrt{T})))},$$
where $K_1$ and $K_2$ are modified Bessel functions of the second kind (with different parameters, 1 and 2).
In this particular setting, preemption is always helpful; 
  in fact $t_1 = \lambda t_2 + 1$. We suggest it is perhaps surprising that these models yield equations with such compact closed forms, albeit in terms of the arguably somewhat obscure modified Bessel functions.
The paper derives other interesting closed forms for cases where predictions are
uniform over $[0,2x]$ for a job of size $x$, and where the service distribution is a particular Weibull distribution.

{\footnotesize
\begin{table*}[!h]
\centering
\begin{tabular}{|c||c||c|c|c||c|c|c|}
\hline
           & FIFO & THRESHOLD  & THRESHOLD   & SRPT & PREDICTION & PREDICTION & SPRPT \\
$\lambda$  &      & NO PREEMPT & PREEMPT &      & NO PREEMPT  & PREEMPT &  \\ [0.5ex]
 \hline\hline
0.50     &      2.000    &      1.783    &      1.564    &      1.425    &      1.850    &      1.698    &      1.659 \\ \hline
0.60     &      2.500    &      2.089    &      1.814    &      1.604    &      2.209    &      2.013    &      1.940 \\ \hline
0.70     &      3.333    &      2.542    &      2.203    &      1.875    &      2.761    &      2.517    &      2.369 \\ \hline
0.80     &      5.000    &      3.329    &      2.910    &      2.355    &      3.757    &      3.451    &      3.143 \\ \hline  
0.90     &      10.00    &      5.278    &      4.755    &      3.552    &      6.366    &      5.960    &      5.097 \\ \hline
0.95     &      20.00    &      8.535    &      7.914    &      5.532    &      10.848   &      10.372   &      8.424 \\ \hline
0.98     &      50.00    &      16.495   &      15.735   &      10.436   &      22.418   &      21.909   &      16.696 \\ \hline
\end{tabular}
\caption{Table from \cite{mitzenmacher2021queues}. Simulation results (except for FIFO) for exponentially distributed service times, using exponential predictions and the optimal threshold.} 
\label{tab:table-exp}
\end{table*}
}

{\footnotesize
\begin{table*}[!h]
\centering
\begin{tabular}{|c||c||c|c|c||c|c|c|}
\hline
 & FIFO & THRESHOLD      & THRESHOLD   & SRPT & PREDICTION & PREDICTION & SPRPT \\
$\lambda$ &      & NO PREEMPT & PREEMPT &      & NO PREEMPT  & PREEMPT &   \\ [0.5ex]
\hline\hline
0.50     &      4.000    &      3.012   &      1.608  &      1.411  &      3.155  &      1.736  &      1.940 \\ \hline
0.60     &      5.500    &      3.676   &      1.867  &      1.574  &      3.918  &      2.062  &      2.280 \\ \hline
0.70     &      8.000    &      4.565   &      2.258  &      1.813  &      4.983  &      2.568  &      2.750 \\ \hline
0.80     &      13.00    &      5.955   &      2.951  &      2.217  &      6.721  &      3.481  &      3.519 \\ \hline
0.90     &      29.00    &      8.940   &      4.649  &      3.154  &      10.630  &      5.790  &      5.224 \\ \hline
0.95     &      58.00    &      13.223  &      7.448  &      4.517  &      16.546  &      9.846  &      7.788 \\ \hline
0.98     &      148.0    &      22.451  &      15.194  &      7.666  &      29.346  &      20.918  &      13.404 \\ \hline
\end{tabular}
\caption{Table from \cite{mitzenmacher2021queues}. Simulation results (except for FIFO) for Weibull distributed service times, using exponential predictions and the optimal threshold.} 
\label{tab:table-ht}
\end{table*}
}

Tables~\ref{tab:table-exp} and \ref{tab:table-ht}, based on data from \cite{mitzenmacher2021queues}, consider simulation results for schemes without predictions and with predictions. The schemes labeled Threshold classify jobs as short or long based on the actual service time (without prediction), showing the impact of having one bit of accurate advice as a comparison point.  They consider the M/G/1 setting with exponentially distributed service times and service times governed by a heavy-tailed Weibull distribution (with cumulative distribution function $F(x) = 1 - e^{-\sqrt{2x}}$).    
Perhaps not surprisingly, 1-bit predictions obtain a large fraction of the benefit of full predictions (which corresponds to SPRPT) in the cases studied. 

Another interesting point is that predictions are even more significant in the context of the heavy-tailed Weibull distribution.  Arguably this is obvious (at least in hindsight) and there is a useful intuition for this. Large queueing delays are caused by longer jobs blocking shorter jobs;  this is why the performance of SRPT can be substantially better than FIFO. Predictions, even if they only get the order mostly right, prevent long jobs from blocking shorter jobs very often.  However, this result provides information back to those designing machine learning models that the right performance metric for 1-bit predictions is not the fraction of correct predictions, because predicting long jobs correctly is more important than predicting short jobs correctly. A mispredicted long job can be placed in front of several shorter jobs, blocking them from service, and significantly increasing all of their times in the system.  A mispredicted short job, however, is only itself hurt when placed at the back of a queue. (See also \cite{dell2015psbs} for discussion of this.) More generally, even with predictions of the actual service times, it is better to have good predictions for longer jobs.   

\medskip

\noindent{{\bf Open Questions:}}
\begin{itemize}
\item What other natural models of prediction errors are there for 1-bit predictions?
\item Can we formalize how to optimize the predictions we would like to obtain from machine-learned predictions, under some specific scheduling policy such as SPRPT or 1-bit predictions from thresholds?
\item 1-bit prediction analysis can readily be generalized to $k$-bit prediction analysis (or from 2 classes to greater than 2 classes).  How do performance characteristics such as the expected response time or the behavior of the tail of the response time vary as $k$ increases?  Note that SOAP-based analyses can also be used to analyze the tail of the response time \cite{scully2020characterizing}.
\end{itemize}

\subsection{Uniform Bounds}

While the ability to derive exact formulae for certain standard queueing models with the addition of prediction is valuable, it can sometimes be difficult to gain more general insights from the specific equations. The analysis methods used for online algorithms, where the input arrives data item by data item, and the algorithm must react as each item arrives, motivates another approach.  Scheduling problems can naturally be seen as online problems, although in the online setting one typically looks at worst-case inputs, rather than stochastic inputs as one generally does in queueing theory.
Many problems in online algorithms have been re-examined in the context of learning-augmented algorithms (see, e.g., \cite{githubio,DBLP:books/cu/20/MitzenmacherV20}), and the two areas fit together quite naturally.  Since in online problems the whole input is not given at the start, achieving an optimal result is not generally possible, and the standard performance measure in online algorithms is the {\em competitive ratio}, which is the ratio of the value of the solution obtained by the proposed online and the value of the optimal solution. (For randomized algorithms, the competitive ratio is usually defined as the ratio between the expected value of the solution obtained by the online algorithm and the optimal.) The question becomes, can the competitive ratio be improved when there is suitable advice?  

In the context of online algorithms with predictions, early work defined two key goals: consistency, which requires near-optimal performance with small error, and robustness, which requires bounded approximation ratio under arbitrary error. 
Formally, as stated in \cite{lykouris2021competitive}, 
we may say that an algorithm is $\alpha$-consistent if its competitive ratio tends to $\alpha$ as the error in the predictions goes to 0, and $\beta$-robust if the competitive ratio is bounded by $\beta$ even with arbitrarily bad predictions.

The work \cite{ScullyGM22} extends these ideas to the setting of M/G/1 queues with predictions.  
The assumption made is that the predictions have bounded multiplicative error, so that a job of size $s$ has predicted size in the range $[\beta s, \alpha s]$ for some $\beta <1 $ and $\alpha > 1$.  (Additionally, the job size and the prediction come from a joint distribution as in the previous work;  that is, the prediction is not chosen adversarially.)  The work first shows that this assumption of bounded multiplicative error is necessary to achieve constant robustness;  that is, without bounded multiplicative error, there are cases where robustness cannot be achieved.  Accordingly, they set a goal of finding a scheduling strategy with the following properties:
\begin{itemize}
\item Consistency: As $\alpha$ and $\beta$ go to 1, the expected response time converges to the expected response time for SRPT (the optimal scheduling algorithm).
\item Graceful Degradation: The ratio of the expected response time of the system using the scheduling strategy and the expected response time of the system using SRPT is bounded by $C \frac{\alpha}{\beta}$ for some constant $C$ and any $\alpha$ and $\beta$.
\end{itemize}
The first goal is a natural form of consistency in this setting.  Graceful degradation, where the performance bound degrades gracefully with the quality of the prediction, appears to be a useful aim in its own right, and is now often considered in works on algorithms with predictions.

Summarizing their work, there are several key points:
\begin{itemize}
\item Simply using SPRPT does not yield expected response times (time in system) bounded within a constant factor of SRPT, even with bounded multiplicative errors.
\item A variant of SPRPT, where the job's rank increases again after reaching 0, is both consistent and provides graceful degradation (with a constant $C$ of 3.5).
\item PSPJF also yields graceful degradation, and the constant $C$ proven for it is in fact better than the constant proven for the SPRPT variant (here $C = 1.5$ is proven).  
\end{itemize}

The analyses utilize ideas from SOAP analysis, along with work integral methods, designed by Scully, Grosof, and Harchol-Balter \cite{scully2020gittins,scully2018soap}.  In particular, there is a careful comparison of rank functions to bound the expected response time of the new SPRPT variant.

\begin{figure}[t]
    \begin{subfigure}[t]{0.31\linewidth}
      \begin{tikzpicture}[figure]
    \yguide[\hphantom{estimated size~$z$}\llap{true size~$s$}]{0}{5.5}
    \xguide[$s$]{5.5}{0}
    \axes{8}{6}{$0$}{age}{$0$}{rank}
    \draw[srpt] (0, 5.5) -- (5.5, 0);
\end{tikzpicture}
    \end{subfigure}
    \begin{subfigure}[t]{0.31\linewidth}
        \begin{tikzpicture}[figure]
    \yguide[estimated size~$z$]{0}{4}
    \xguide[$z$]{4}{0}
    \axes{8}{6}{$0$}{age}{$0$}{rank}
    \draw[srpt-e] (0, 4) -- (5.5, -1.5);
\end{tikzpicture}
    \end{subfigure}
    \begin{subfigure}[t]{0.31\linewidth}
      \begin{tikzpicture}[figure]
    \yguide[estimated size~$z$]{8}{4}
    \xguide[$z$]{4}{0}
    \axes{8}{6}{$0$}{age}{$0$}{rank}
    \draw[srpt-b] (0, 4) -- (4, 0) -- (8, 4) -- (10, 4);
\end{tikzpicture}
    \end{subfigure}
    \caption{Rank functions of size-estimate-based policies.  The rank function for SRPT is on the left;  the rank decreases as $s-a$ where $s$ is the true size and $a$ is the age.  The rank function for SPRPT is in the middle; 
    it decreases as $z-a$ where $z$ is now the estimated size.
    Note that a job can have negative rank, at which point it cannot be preempted. The SPRPT-with-bounce rank function from \cite{ScullyGM22} is on the right; the rank decreases from the estimate $z$ to 0 but bounces back up, according to the function $\max(|z-a|,z)$. This rank bounce tempers the effect of long jobs that are predicted to be short delaying short jobs from being served.}
    \label{fig:rank}
\end{figure}
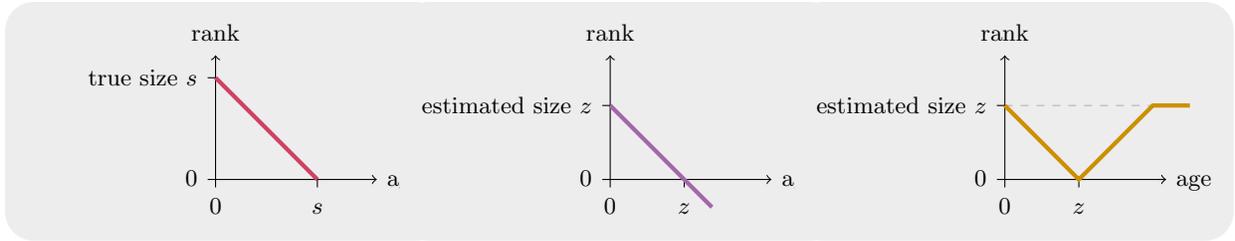

It is worth noting that the more traditional, worst-case scheduling problems have similarly been studied as online algorithms problems.  Most similarly, Azar, Leonardi, and Touitou examine the classic online problem of scheduling on a single machine to minimize total flow time, when job times may be distorted by up to a multiplicative factor of $\mu$. Their first work \cite{DBLP:conf/stoc/AzarLT21} shows that
for every distortion factor $\mu$, there is an $O(\mu^2)$-competitive algorithm, but the algorithm needs to know $\mu$ in advance.  In later work \cite{DBLP:conf/soda/AzarLT22}, they improve this to provide a specific $O(\mu \log \mu)$-competitive algorithm that does not know $\mu$ in advance. These works therefore obtain similar results to \cite{ScullyGM22}, without making stochastic assumptions on the job sizes, but with a more complex algorithm and a slightly larger-than-linear competitive ratio.  

\medskip

\noindent{{\bf Open Questions:}}
\begin{itemize}
\item Can we tighten the various bounds of \cite{ScullyGM22}, in particular improving the constants $C$ related to graceful degradation?
\item Could more complex scheduling algorithms, going beyond rank-based algorithms, yield better performance bounds?  
\item The results focus on the expected response time. Are there other important performance measures, and how should they be evaluated in this setting?
\end{itemize}

\subsection{Accounting for Prediction Costs}

The results presented above demonstrate the great potential for using predictions to improve scheduling performance. However, the models we have discussed all suffer a somewhat glaring flaw: they do not model the resources required to obtain such predictions, but instead assume that predictions are provided ``for free'' when a job arrives. To be fair, this assumption is not unusual in the area of algorithms with predictions more generally, as the prediction cost may be small in the context of the algorithm, or the focus may be on a different performance metric. For scheduling in particular, however, 
assuming prediction costs can be ignored may not be realistic, as the resources devoted to calculating predictions might be more effectively used to directly serve the jobs themselves. This perspective challenges the potential effectiveness of integrating predictions into real-world queueing systems.

This point is considered in \cite{DBLP:conf/nips/ShahoutM24}, which incorporates costs into the analysis of M/G/1 queues with predictions, and considers novel scheduling approaches that take these costs into account. In that work, they consider two models of costs. In the first model, referred to as the external cost model, predictions are provided by some external process and do not affect job service time, but there is a fixed cost for predictions. The expected cost per job in this model would naturally be taken as the sum of the job’s expected response time within the system and the prediction costs.  With this model, one might
consider only using predictions for some jobs but not others.
In the second model, referred to as the server time cost model, predictions themselves require a fixed time from the same server that is servicing the jobs, and hence a scheduling policy involves also scheduling the predictions.
The expected cost per job in this model is just the expected response time.
Note that, because the predictions require work from the server, there are more complex interactions; in particular, for heavily loaded systems, the time used for jobs to obtain predictions could lead to an overloaded, unstable system.

\begin{figure}[H]
    \centering
    \includegraphics[width=0.58\linewidth]{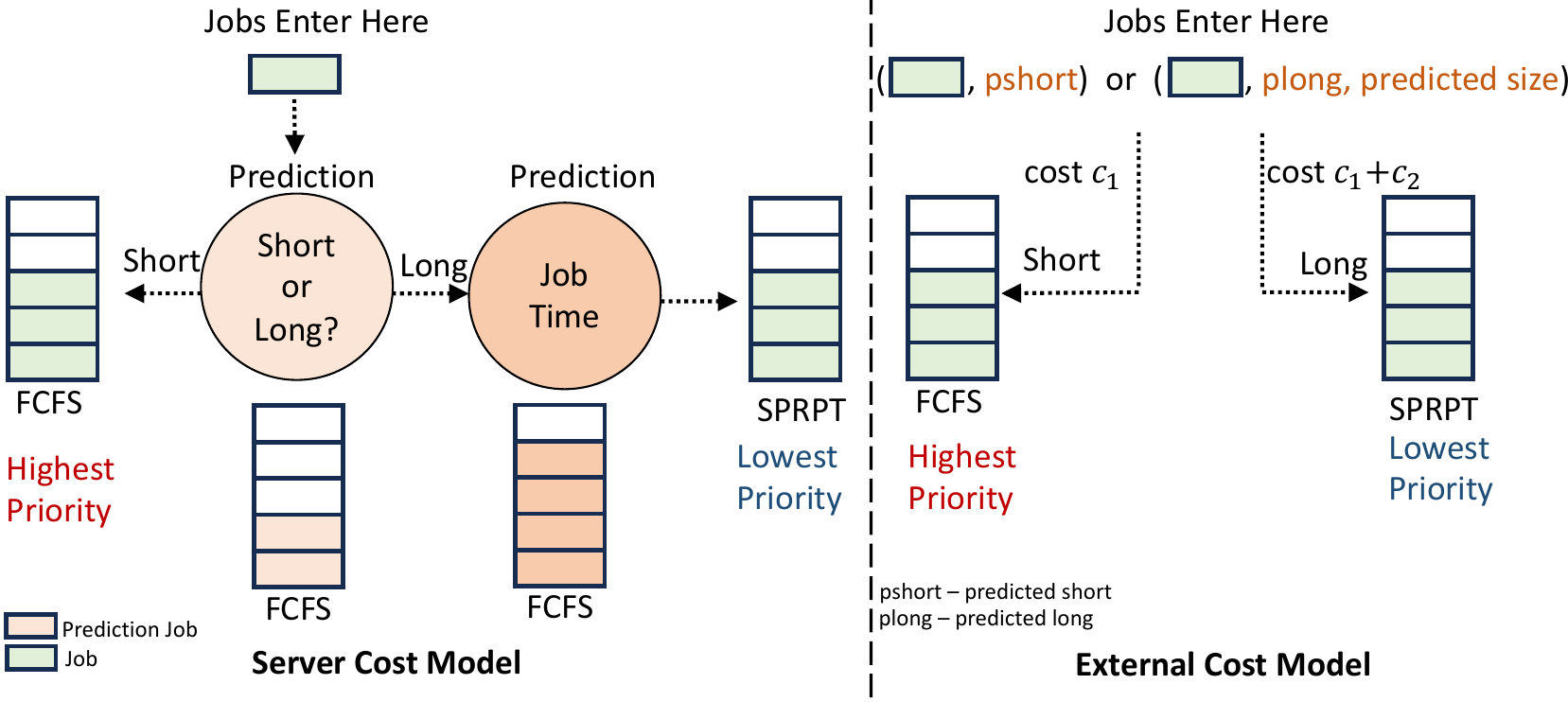}
    \caption{SkipPredict framework under the server cost model and external cost model.}
    \label{fig:skippredict}
\end{figure}

As a starting point, \cite{DBLP:conf/nips/ShahoutM24} derives the formulas for SPRPT and 1-bit predictions under both cost models, which can again be done using a SOAP analysis \cite{scully2018soap}. However, \cite{DBLP:conf/nips/ShahoutM24} argues that the introduction of costs allows for more interesting models and scheduling strategies. They focus on a setting where 1-bit predictions are cheap compared to a prediction of the service time. In such a setting, it may make sense to use the cheap 1-bit prediction for all jobs, but only use the more expensive prediction for long jobs. Predicted short jobs are scheduled by FIFO, and predicted long jobs are scheduled after short jobs, using SPRPT.  They call this scheduling algorithm SkipPredict (Figure~\ref{fig:skippredict}), and show it also can be analyzed using SOAP analysis.

The SOAP analyses provided are somewhat complex, as they utilize two-dimensional ranking functions to provide priorities.  For example, for SkipPredict in the server cost model, short jobs always have highest priority, followed by 1-bit predictions for jobs that have entered and not received such predictions.  Then, priority goes to predictions for long jobs, and long jobs themselves have the lowest priority. This prevents known short jobs from being delayed by other jobs, and ensures that long jobs are handled by SPRPT when long jobs are being serviced.  

\begin{figure}[H]
    \centering
    \includegraphics[width=0.58\linewidth]{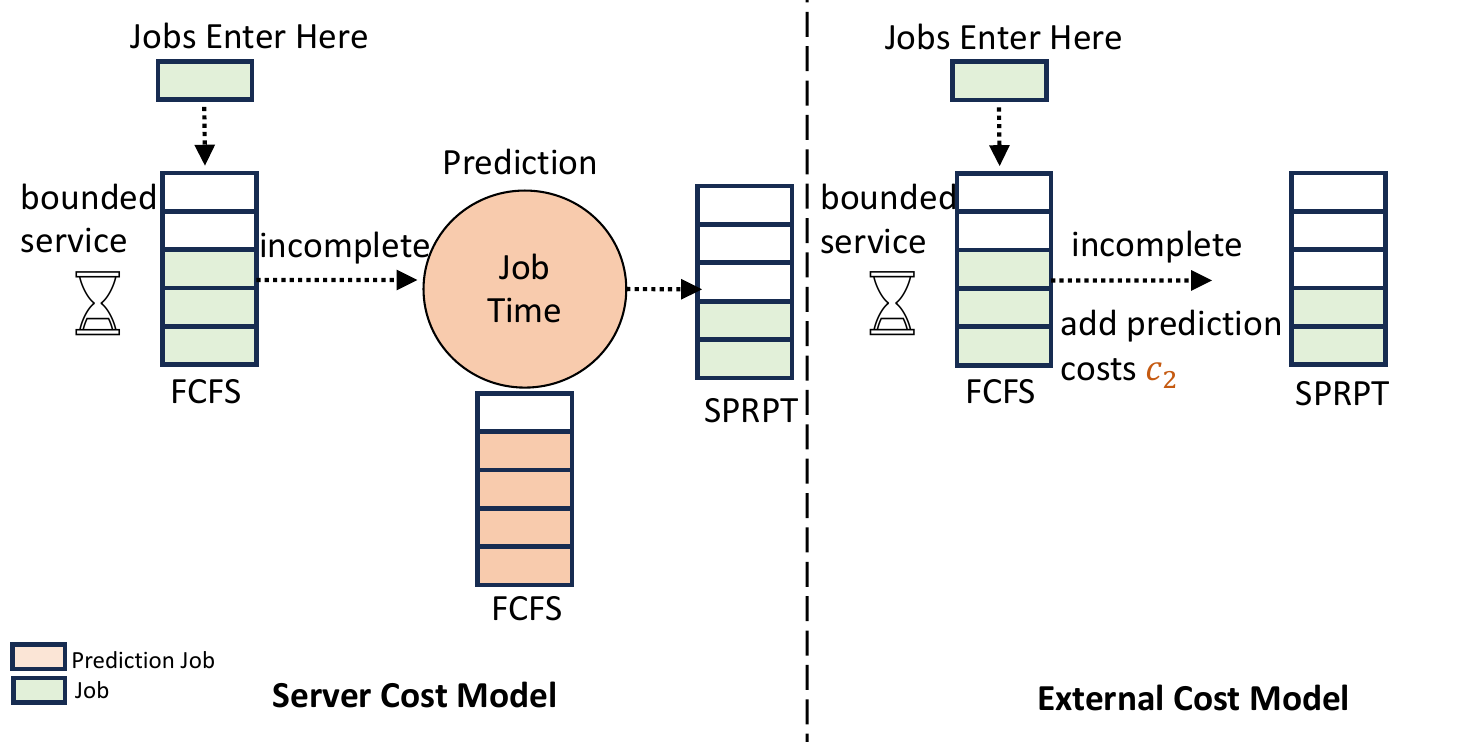}
    \caption{DelayPredict framework under the server cost model and external cost model.}
    \label{fig:delaypredict}
\end{figure}

As another alternative, they analyze
a scheduling algorithm, DelayPredict, that avoids cheap predictions entirely, but still limits the jobs that undergo more expensive predictions of the service time.
DelayPredict initially schedules all jobs in a FIFO manner, but instead of using cheap predictions, it limits each job to a limit $L$ of time, at which point the job is preempted and treated as a long job. At that point, the job will be deprioritized and queued for prediction;  longer jobs are then served by SPRPT. DelayPredict provides an alternative to SPRPT that avoids the cost of predictions for every job, and can lead to improvements if 1-bit predictions are either not much cheaper than full predictions, or not available. DelayPredict framework is shown in Figure~\ref{fig:delaypredict}.

\medskip

\noindent{{\bf Open Questions:}}
\begin{itemize}
\item Are there other natural models of prediction costs for queueing systems? 
For example, rather than having a separate prediction stage, one could obtain a prediction as the job runs, at some cost (slowdown) of the job for some initial time period.  
\item When predictions have cost, they may not be worthwhile when the system is heavily loaded. Can we design and analyze scheduling policies that choose when to use predictions based on the current load, or otherwise respond dynamically in choosing when to use predictions?
\item How can selective prediction strategies be integrated into scheduling to balance prediction cost and performance? Specifically, can we design a scheme where only a subset of jobs is predicted (e.g., with some probability), while jobs without predictions are handled via FIFO?
\item Can we design systems that use dedicated prediction servers effectively?

\item In \cite{grosof2022incentive}, a different notion similar to cost is considered; jobs with incorrect predictions are possibly punished by being assigned lower priority. This model is used to examine incentive compatibility in the context of self-reported predictions.
Are there other ways of adding economic considerations to refine prediction models?
\end{itemize}

\subsection{Multiple Server Systems}

The above work has focused on using predictions in the setting of the M/G/1 queue. Using predictions in larger systems or networks of queues remains relatively open for further study. Most previous work has been empirical \cite{dell2019scheduling,dell2015psbs,mailach2017scheduling}, such as the work \cite{mitzenmacherd2022supermarket} which looks at the power of two choice paradigm \cite{mitzenmacher2001power,Suhov1997} when using predictions.  However, recent work by Dong and Ibrahim has examined the asymptotic performance of shortest predicted job first with noisy estimates of job sizes for 
M/GI/$s+$GI systems, showing that it asymptotically maximizes the throughput 
among all non-preemptive scheduling disciplines that use that noisy service-time information \cite{dong2024shortest}.
The lack of theoretical work thus far is arguably not surprising;  multiple server systems resist analysis even without predictions, and there are many open questions for systems with multiple servers that become only more challenging when adding predictions.  (For recent work on multiple server systems generally see, e.g., \cite{grosof2024optimal}.)

Another small step forward appears in the work on 1-bit predictions \cite{mitzenmacher2021queues}, where a variant of the power of two choices is considered in the setting of 1-bit predictions by deriving the fluid limit differential equations.  In this setting, each job chooses $d$ queues uniformly at random, and the job chooses to wait at the best of the $d$ choices for some defined notion of best.  (In \cite{mitzenmacher2021queues}, the decision is based on the number of jobs of the same predicted type that are queued.)  The fluid limit system corresponds to the number of queues going to infinity.
This analysis requires Poisson arrivals and exponentially distributed service times for the long and short jobs.  The key to this analysis is that the set of queue states has a short description: the state can be represented by the number of queued jobs that are predicted to be short and long, and whether the current running job is short or long. 
\cite{mitzenmacher2021queues} also provides an interesting example where the prediction error rate is greater than 50\% over all jobs, but using predictions still performs better than not using predictions in the fluid limit.  
As one might expect, in this example the prediction error rate is made large for short jobs, and smaller for long jobs, showing the importance of predicting long jobs accurately.

\medskip
\noindent{{\bf Open Questions:}}
\begin{itemize}
\item Can we generalize any existing theoretical work on multiple server systems to systems with predictions naturally?
\item In particular, can we develop fluid limit models to analyze the power of two choices when using (more than 1-bit) predictions?
\item Another area where predictions may be useful is in multiserver-job systems, where jobs may run concurrently on many servers. As a challenging example, consider a setting where jobs may use a variable number of servers with the running time dependent on the number of servers used, and there are predicted times associated with each possible number of servers the job could use.
That is, one is predicting the speedup obtainable for a job by using more servers, which may vary depending on the job type.
\end{itemize}

\section{Large Language Model Systems}
\label{sec:llm_background}


LLMs have revolutionized artificial intelligence by moving beyond predicting tokens to solving adaptive problems. Their impact spans across professional sectors, from healthcare and finance to education and customer support, where they enable decision support and personalized interactions. In the creative and technical domains, LLMs can generate artistic content~\cite{ramesh2021zero}, automate code development~\cite{chen2021evaluating}, and accelerate scientific research through data analysis and literature synthesis \cite{jo2023promise}. ChatGPT~\cite{achiam2023gpt} exemplifies how LLMs can enhance AI capabilities through natural dialogue, enabling users to engage with AI systems for tasks ranging from simple queries to complex problem-solving. 

In high-concurrency environments, users expect real-time responses, which makes minimizing latency essential for a seamless experience. Scheduling can address this latency optimization challenge by minimizing the average user response time (the time from when a request first arrives until it completes service) or by affecting the tail of the response time distribution.
The scheduler decides which requests to queue or serve and how to order the requests within each queue. (Since many applications rely on pre-trained models for inference instead of training their own, we focus on scheduling during inference, although scheduling considerations may also apply to training LLMs.)
With respect to minimizing use response time, scheduling in LLMs feels similar to queueing theory models like those we have discussed.
However, with LLMs
the scheduler might have additional goals to optimize for, such as throughput and cost, introducing further challenges.

Predictions can naturally help inform scheduling decisions for LLMs, as many job details may be unknown on arrival. Ideally, these predictions are made before running a request, but they can also be refined during execution as the system gathers insights from the LLM itself. Also, as we discuss further later, predictions in the LLM setting can extend beyond estimating request sizes.

To explain how queueing models apply to LLM systems, we first provide background on LLM inference and hardware execution. We then discuss how LLM systems differ from traditional queueing systems and the challenges these differences introduce.
We believe both that the research community needs to develop new queueing theory models tailored to LLM systems, and that existing queueing theory can provide both analysis methods and heuristic insights that can improve LLM scheduling. 

\subsection{LLM Inference}
LLM models follow an autoregressive pattern, where text is generated one token (i.e. one or more words or parts of words) at a time, based on the calculated probability distribution for the next token given the preceding context. This process, referred to as inference, consists of sequential iterations where each iteration generates a token and appends it to the existing input prompt. The generation continues until a termination condition, such as a predefined maximum output length or an end-of-sequence token, is met.

\subsubsection*{Transformer Architecture}

\begin{figure}[h]
    \centering
    \includegraphics[width=0.75\linewidth]{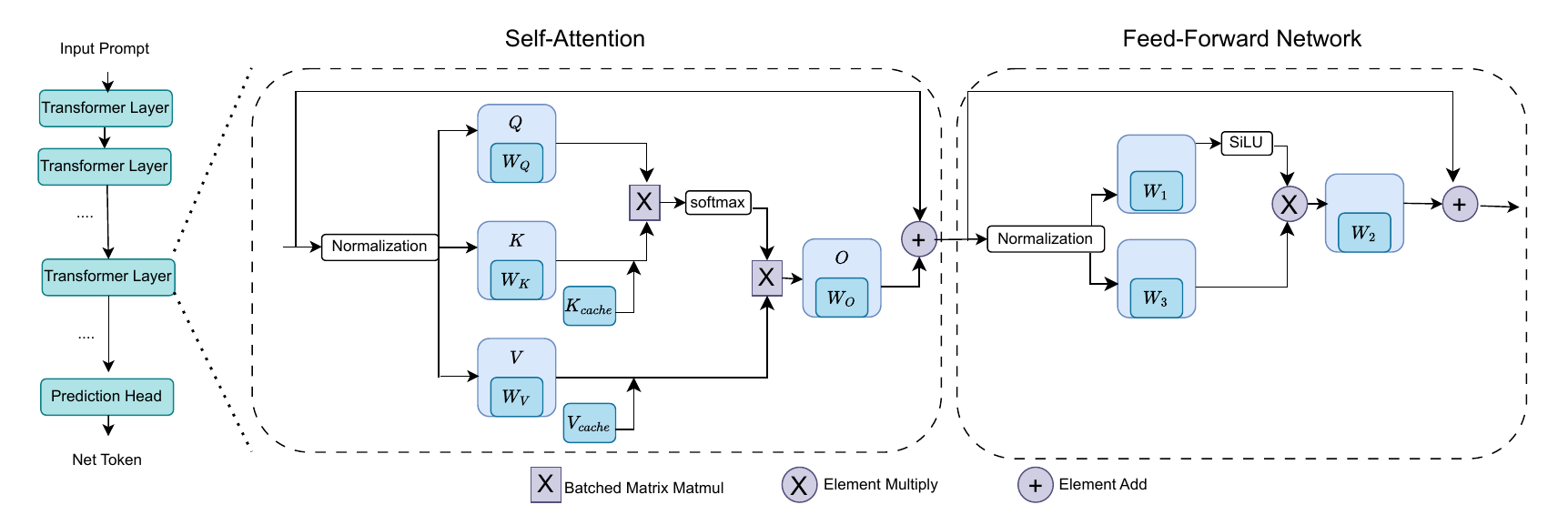}
    \caption{Transformer architecture}
    \label{fig:transformer_arch}
\end{figure}

Most of today’s LLMs adopt a decoder only transformer
architecture~\cite{brown2020language, radford2019language}.
The input to a transformer layer is an embedding of the tokenized input sequence. Each transformer model consists of a stack of sequential layers, where each layer applies self-attention mechanisms to capture contextual relationships between tokens and feed-forward networks to refine the representation of the input as shown in Figure~\ref{fig:transformer_arch}.



\paragraph*{Self-attention mechanism.}
The self-attention mechanism enables the model to weigh the importance of different tokens in the input sequence when making predictions. For a given input \( X_{\text{pre}} \in \mathbb{R}^{n \times d} \), where \( n \) is the sequence length and \( d \) is the hidden dimension (the embedding size of each token), the model applies learned linear transformations to produce the query (\( Q_{\text{pre}} \)), key (\( K_{\text{pre}} \)), and value (\( V_{\text{pre}} \)) matrices:

$$
Q_{\text{pre}} = X_{\text{pre}} W_q, \quad K_{\text{pre}} = X_{\text{pre}} W_k, \quad V_{\text{pre}} = X_{\text{pre}} W_v
$$
where \( W_q, W_k, W_v \in \mathbb{R}^{d \times d_k} \) are learnable weight matrices that project the input embeddings into a lower-dimensional space of size \( d_k \) that represents the number of columns in \( K_{\text{pre}} \), determines the size of the query-key dot product.

These queries, keys, and values are used to compute the attention output through the following formula:

$$
O_{\text{pre}} = \text{softmax} \left( \frac{Q_{\text{pre}} K_{\text{pre}}^T}{\sqrt{d_k}} \right) V_{\text{pre}} W_o + X_{\text{pre}}
\label{eq:prefill_eq}
$$
where \( W_q \), \( W_k \), \( W_v \), and \( W_o \) are the learnable weight matrices. The softmax function ensures that attention weights sum to one across each row, allowing the model to assign different importance levels to tokens. It is defined as: $\text{softmax}(z_i) = \frac{\exp(z_i)}{\sum_j \exp(z_j)}$, where each $z_i$ represents an element of the input matrix.


\paragraph*{Feed forward network.}

The output of the self-attention module is sent to the Feed-Forward Network (FFN), which refines the attention output. This network introduces non-linearity, allowing the model to capture more complex patterns in the data.

$$
\text{FFN}(x) = \left( \text{SiLU}\left(x W_1\right) \times x W_3 \right)  W_2,
$$
where \( W_1 \), \( W_2 \), \( W_3 \) are linear modules. The SiLU (Sigmoid Linear Unit) activation function is defined as: $\text{SiLU}(x) = x \cdot \sigma(x) = x \cdot \frac{1}{1 + e^{-x}}$, where $\sigma(x)$ is the sigmoid function, defined as $\frac{1}{1 + e^{-x}}$.


\paragraph*{Prefill and decode phases.}

LLM inference is divided into two phases, \emph{prefill} and \emph{decode}. The prefill phase is the initial step, where the model processes the input prompt ($X_{\text{pre}}$) and generates key-value pairs that are stored in the KV cache, which holds contextual information required for generating subsequent tokens. The design of the transformer allows for parallel processing of input tokens during the prefill phase. In the decode phase, new tokens are generated based on previous tokens, step by step. The input of this phase is \( X_{\text{dec}} \in \mathbb{R}^{1 \times d} \) and the model retrieves previously stored key-value pairs from the KV cache to continue generating tokens where after each generated token, new key-value pairs are computed and appended to the existing cache.

$$
Q_{\text{dec}} = X_{\text{dec}} W_q, \quad K_{\text{cat}} = [K_{\text{cache}}, X_{\text{dec}} W_k], \quad V_{\text{cat}} = [V_{\text{cache}}, X_{\text{dec}} W_v]
$$

The attention output in the decode phase is computed as:

$$
O_{\text{dec}} = \text{softmax} \left( \frac{Q_{\text{dec}} K_{\text{cat}}^T}{\sqrt{d_k}} \right) V_{\text{cat}} W_o + X_{\text{dec}}
$$


Storing and loading the intermediate attention matrix increases inference time in transformer architectures. FlashAttention~\cite{dao2022flashattention, dao2023flashattention} and Flash-Decoding~\cite{dao2023flash} combine matrix multiplications and the softmax operator in self-attention into a single operator. This integration eliminates the need to store and load the intermediate matrix, reducing both memory access overhead and inference time.
The KV cache grows over the iterations for inference, introducing unique challenges in optimizing latency -- a consideration specific to inference, as the KV cache is not present during training.

\subsection*{Execution on Hardware}

\begin{figure}[h]
    \centering
    \includegraphics[width=0.6\linewidth]{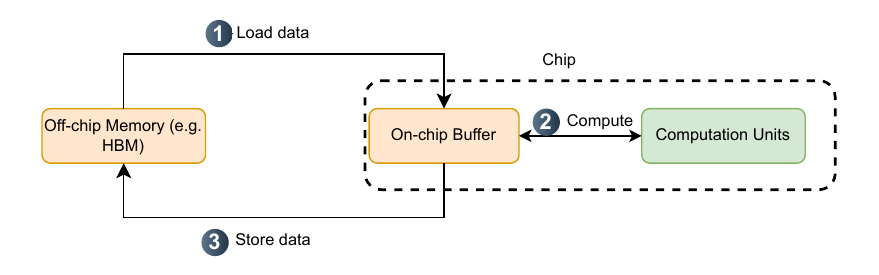}
    \caption{A neural network layer is executed on hardware devices by transferring data from memory (e.g., HBM) to on-chip buffers, then computing with the on-chip processing units, and eventually sending the output data back to memory.}
    \label{fig:execution}
\end{figure}

The rapid progress in LLMs is closely tied to advancements in hardware accelerators, particularly GPUs (Graphics Processing Units). Unlike traditional CPUs, which execute tasks sequentially, GPUs are optimized for parallel processing. They consist of thousands of small cores capable of performing parallel computations, making GPUs particularly effective for matrix and vector operations that are fundamental to transformer-based models. A key feature that enhances GPU performance is the use of high-bandwidth memory (HBM), which enables faster data transfer between memory and processing units. As illustrated in Figure~\ref{fig:execution}, executing a neural network layer on GPU involves three steps: transferring data (e.g. model weights and KV cache) from memory (such as HBM) to on-chip buffers, performing computations within the on-chip processing units, and writing the results back to memory. The efficiency of this process is influenced by both memory access speed and the computational capacity of the processing units.

The imbalance between computation and memory access may, however, lead to performance bottlenecks. When a layer involves significant computations but minimal memory access, it creates a \textit{computation bound}, leaving memory units idle while computations are processed. Alternatively, a \textit{memory-bandwidth bound} arises when a layer demands extensive memory access but performs fewer computational tasks, resulting in underutilized GPU processing cores.

The prefill and decoding phases differ in their use of computation and memory. The prefill phase efficiently uses GPU parallelism, as each input is processed independently and all inputs are available upfront. This makes the prefill phase compute bound. In contrast, the decode phase is memory-bandwidth bound. During decoding, significant GPU memory bandwidth is used to load model parameters, often making data transfers to the compute cores slower than the actual token processing. Batching multiple requests during the decode phase helps to reduce this bottleneck by loading model parameters once and reusing them for multiple inputs, which increases throughput and reduces inference costs. Thus, LLM inference throughput depends heavily on the number of requests that can be batched into the GPU's high-bandwidth memory.

\subsection{Performance Metrics}
LLM systems extend traditional performance metrics while introducing new considerations. 
Although model size strongly influences performance, it is not the only factor. Key metrics include:

\begin{itemize}
    \item \textbf{Computational cost:} Depends on the model’s size (its parameter 
    count) and the length of the generated response, especially given the autoregressive nature 
    of LLM inference.
    \item \textbf{Latency (or response time):} The time from request arrival until service completion. Latency can be measured in two ways: (a) overall latency, which is affected by model size as well as prompt and output lengths, and (b) time-to-first-token (TTFT), which is the time from request arrival until the first token is produced, primarily influenced by model size and prompt length.
    \item \textbf{Throughput:} The number of tokens generated per second.
    \item \textbf{Accuracy:} While larger models often produce higher-quality responses, the 
    link between model size and accuracy varies with the request type.
    \item \textbf{Energy (or carbon footprint):} The energy consumed and the resulting carbon emissions during LLM inference, typically measured in kilowatt-hours (kWh) for energy and grams of CO2 per request or per token. Key factors include hardware specifications (e.g., GPU type), model size, batch size, and data center location. LLMCarbon~\cite{faiz2023llmcarbon} offer estimates of LLM carbon footprints.
\end{itemize}

We focus on the question of reducing overall latency (which we refer to henceforth as just latency). Although using a smaller model can lower latency, it may compromise accuracy. Our aim is to develop scheduling policies that minimize response times without altering the chosen LLM model, which we take as a given. Problems where model selection or modification enables an accuracy-latency tradeoff are an interesting direction mentioned in Section~\ref{sec:compound_systems}.

We define a request’s latency as the time from when a user submits the request until the answer is returned. In an LLM system, two additional metrics are used to evaluate the system's performance in handling requests: Time To First Token (TTFT) measures how long it takes to generate the first token, which often depends on the prompt length (i.e., the prefill phase). Time Per Output Token (TPOT) measures the time required to generate each subsequent token (i.e., one decode phase). 
For a request $R$, with an input size of $n_\text{input}$ tokens, an output size of $n_\text{output}$ tokens, and a waiting time $t_\text{waiting}$, the latency of $R$ is defined as:
\begin{align}
t_\text{response} = t_{waiting} + \mathrm{TTFT}(n_\text{input}) + n_\text{output} \cdot \mathrm{TPOT}.
\label{eq:latency}
\end{align}

\subsection{A Summary: The Job vs. System Perspective}

Before going into details regarding LLM scheduling, we summarize the operations we have described, from two perspectives: the job\footnote{The terms job and request are used interchangeably in this paper.} perspective and the system perspective.

\textbf{Job perspective:} This perspective examines individual jobs by analyzing their journey through processing phases, including the time spent in the queue. Each inference job consists of an input (prompt) and an output, both measured in tokens for convenience. A job undergoes two phases: prefill and decode. During the prefill phase, the model applies learned linear transformations to the input to produce information for the KV cache, with memory usage proportional to the input size. This phase executes as a single computational block because the entire prompt is available. In the decode phase, the model generates tokens sequentially in an autoregressive manner; with each iteration, a new token is produced and an additional KV cache entry is added, causing the KV cache to grow in proportion to the combined input and output sizes. Preemption in LLM systems is at the token level~\cite{yu2022orca}; however, preemption poses challenges because the KV cache must be retained until the request is fully processed, or else previously computed work is lost and must be recomputed. Alternatively, a job may be partially terminated by discarding the most recent (tail) portion of the KV cache.  

\textbf{System perspective:} This perspective considers the management of jobs at the system level through batching, where multiple jobs are served simultaneously. Batching allows the system to load model parameters once and reuse them for multiple jobs, optimizing resource utilization. With continuous batching, introduced by Orca~\cite{yu2022orca}, where new requests can join an existing batch and completed requests are returned immediately at the iteration level rather than waiting for an entire batch to finish; a single batch may comprise jobs in different processing phases (prefill and decode). Batching is performed at the token level: at each token time unit, the system forms a batch of jobs, some in the prefill phase and others in the decode phase, and processes them concurrently via processor sharing, using all available computational resources. After all jobs in a batch complete (so either the prefill completes, or the decode generates a new token for the job), the system updates the batch by removing completed jobs, adding new ones, and possibly preempting existing jobs, with the goal of ensuring continuous resource reallocation to maximize performance. Building on this, vLLM~\cite{kwon2023efficient} integrates paged attention which allocates the KV cache gradually in blocks during inference instead of allocating for the maximum output length at the beginning, resulting in improved throughput and reduced operational costs. vLLM improves the throughput of popular LLMs by $2-4x$ with the same level of latency compared to Orca and FasterTrasnformer~\cite{fastertransformer}. vLLM has been widely adopted and established itself as the state-of-the-art framework for inference services.

Finally, we remark that in typical transformer architectures, processing occurs sequentially across neural network layers during both the prefill and decode phases. For scheduling purposes, we treat all layers as a single block and do not consider intra-layer scheduling, although exploring this finer granularity remains an interesting possible direction.

\subsection{How Do LLM Serving Systems Differ from Standard Queueing Systems?}
\label{sec:challenges}

Standard scheduling approaches often fall short for LLM serving because LLM systems present unique characteristics and challenges not found in typical systems. We summarize some of these key issues.

\paragraph{KV cache during inference.}
During inference, LLMs generate a KV cache that remains in memory until the request is completed. This contrasts with standard queueing systems, where jobs do not maintain a large memory footprint throughout processing, and in particular do not consider memory requirements that grow linearly with the request length. The KV cache reduces computation time but demands substantial memory, proportional to the model’s number of layers and hidden dimensions. For instance, a single GPT-3 175B request with a sequence length of 512 tokens requires about 2.3 GB of memory for key-value pairs.

\paragraph{Preemption overhead.}
Preemptive scheduling is essential in online settings, where new requests arrive during the execution of longer requests. 
From a job perspective, processing a single request simplifies KV cache management because there are no competing jobs for memory, even when preemption occurs. In contrast, from a system perspective, batch processing requires careful scheduling to manage the KV cache across multiple simultaneous jobs, and preemptions must be limited to ensure that each job completes without triggering memory overflow.

Given the large memory footprint of LLMs and the limited GPU capacity, this overhead can lead to memory exhaustion. Non-preemptive policies, such as FCFS, avoid this overhead but often result in higher response times.
One approach to mitigate this overhead suggests that inactive KV tensors could be offloaded to CPU memory and reloaded into GPU memory when needed. However, the overhead of offloading and reloading is nontrivial compared to token generation time. For example, deploying GPT-3 175B on NVIDIA A100 GPUs requires approximately 2.3 GB of memory per job for KV tensors. During decoding, token generation takes about 250 ms, whereas transferring KV tensors between host and GPU memory over PCIe 4.0×16 at full bandwidth takes about 36 ms. Existing approaches~\cite{abhyankarinfercept, wu2023fast} attempt to optimize offloading and reloading by overlapping these operations with computation. However, the available memory budget for such overlap poses a fundamental constraint.

\paragraph{Multi-stage processing.}
Requests in LLM systems can be viewed as multi-stage jobs in queueing theory, though they differ from standard queueing models. Mixed-phase batches, where some requests remain in the prefill phase while others have advanced to decoding, can prolong the overall decode phase since the longer prefill phase may dominate the iteration time.
Two strategies have been proposed to mitigate this imbalance. The first, \emph{chunked prefill} (Sarathi~\cite{agrawal2024taming}), splits prompt tokens into smaller segments that are processed alongside decode requests in each batch iteration. In this abstraction, a request is treated as having two parts with distinct processing times. The second strategy, \emph{split phases} (Splitwise~\cite{patel2024splitwise, zhong2024distserve}), separates the prefill and decode phases across different machines, aligning processing resources with the computational demands of each phase. In split phases, different GPUs handle the prefill and decode phases, each operating with its own processing rate and memory constraints. The prefill machine transfers the KV cache to the decode machine, introducing a transfer delay that adds overhead. To mitigate this issue, \cite{patel2024splitwise} optimizes KV cache transfers by leveraging high-speed Infiniband interconnects. This arrangement differs from assuming that all servers have uniform capabilities and can process any job interchangeably. Moreover, distributing these phases across multiple machines requires scheduling decisions regarding where to process each part of a request.

\section{Scheduling in LLM Serving}
\label{sec:llm_scheduling}

We consider in this section a single LLM deployment that spans one or more GPUs, depending on the model's size. When the model's memory footprint requires multiple GPUs, we simplify the scheduling problem by treating the distributed model as a single unit, abstracting away the inter-GPU communication overhead.

\subsection{Dynamic Batching and Preemption in LLM Inference}
\begin{figure}[ht]
    \centering
    \includegraphics[width=0.65\linewidth]{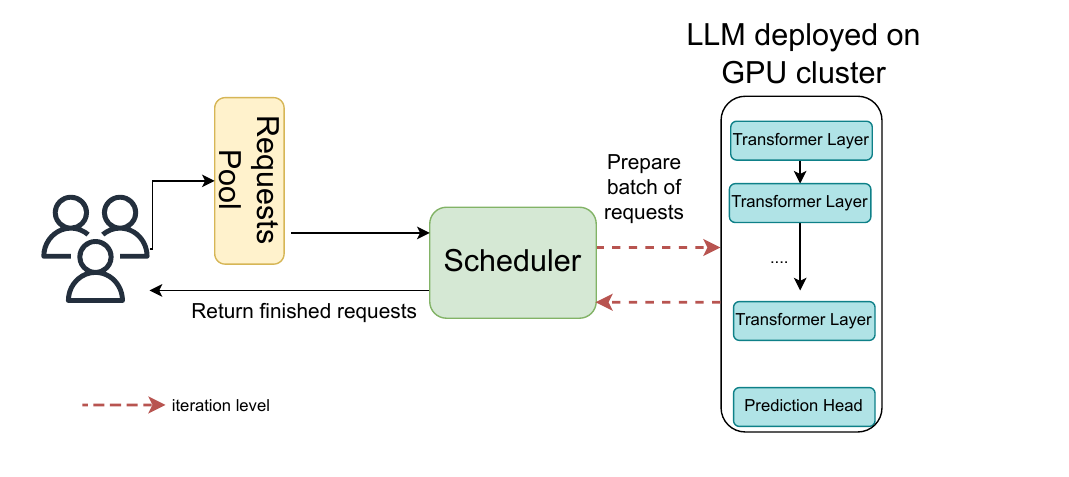}
    \vspace{5pt} 
    \caption{
        \textbf{Iteration-level scheduling.} The scheduler dynamically adjusts the batch and enables preemption decisions. After each token generation, it evaluates whether to continue with the current request or switch to another pending request. Finished outputs are returned to users as soon as they are ready.
    }
    \label{fig:iteration_scheduling}
\end{figure}

LLM inference systems commonly employ iteration-level scheduling~\cite{yu2022orca} (Figure~\ref{fig:iteration_scheduling}), also known as continuous or dynamic batching. Unlike traditional request-level scheduling, where a fixed batch runs to completion before processing a new batch, iteration-level scheduling operates token by token.
This approach allows the scheduler to return finished requests to the user and to adjust the batch after each iteration and enables preemptive scheduling at the granularity of individual tokens. After each token is generated, the scheduler evaluates whether to continue processing the current request or switch to another pending request.

Request demands vary: some requests have a short yes/no output, while others seek lengthy explanations. This variability poses a challenge because short requests may wait behind larger ones, resulting in longer response times. 
From a mean response time perspective, it is more efficient to give short requests priority, as in policies like Shortest Remaining Processing Time. However, implementing SRPT in LLM systems poses unique challenges. We typically need accurate request sizes to use size-based scheduling, yet these sizes are often unknown. In LLM systems, a request's size depends on both the input size (in tokens) and the output size, and the latter is not known a priori. Predicting output size from a prompt is challenging because LLMs generate text autoregressively: each generated token is appended to the input, dynamically altering the context for subsequent token generation.


Several works propose methods to predict request sizes. 
Zhen et al.~\cite{zheng2024response} use an auxiliary LLM model to predict response sizes and then prioritize requests based on those predictions. While this strategy reduces response time, it introduces additional computational overhead from the extra model used for size prediction.
$S^3$~\cite{jin2023s} fine-tunes a BERT model~\cite{sanh2019distilbert} to predict output sequence sizes from input prompts. The studies in \cite{cheng2024enabling,jin2023s, stojkovic2024dynamollm} address size prediction as a classification task, where predictions correspond to one of several buckets representing a range of sizes, whereas \cite{qiu2024power,qiu2024efficient}, use regression-based approaches to estimate a specific size. 
Although these prediction models are relatively lightweight, their accuracy declines for requests with highly variable execution times.
LTR~\cite{fu2024efficient} employs a Learning-to-Rank approach. Instead of predicting the absolute output size of a request, LTR ranks requests based on their output size, allowing the system to prioritize those with fewer remaining tokens. Although ranking is a simpler task than absolute size prediction, it requires training a ranking model in an offline phase.
A limitation of LTR is that it ignores the size of the prompt when ranking, considering only the output size. This absence can lead to head-of-line blocking, particularly when a request with a short output is preceded by a lengthy prompt during the prefill phase.
Given the difficulties in predicting output sizes for LLMs, 
FastServe~\cite{wu2023fast} is based on Multi-Level Feedback Queue (MLFQ) scheduling where each job starts in the highest priority queue and is downgraded if it exceeds a set threshold, resulting in frequent preemptions. To counter this, FastServe introduces a proactive GPU memory management mechanism that offloads and uploads intermediate states between GPU and CPU.
Compared to Orca, FastServe improves the average and tail job completion time by up to $5.1 \times$ and $6.4 \times$, respectively.

Preemptive scheduling can further reduce latency in online LLM services, where new requests may arrive during the execution of longer ones. By interrupting a long-running request to serve a shorter one, overall latency is reduced. Yet, a key complication arises from the need to retain the KV cache for any preempted request, consuming scarce memory resources (see Section~\ref{sec:challenges}).

Trail~\cite{trail} addresses both output size prediction and the preemption overhead. For output size prediction, Trail uses the autoregressive nature of LLM output generation. It recycles embeddings from intermediate transformer layers and processes them with a lightweight linear classifier to estimate the remaining output size. This approach, known as \emph{probing}~\cite{belinkov2022probing, hewitt2019designing, hewitt2019structural}, combines the benefits of direct LLM-based predictions with computational efficiency, eliminating the need for a separate size-prediction model.
To tackle the preemption challenge, Trail proposes a variant of SPRPT. In standard SPRPT, a newly arrived request with a shorter predicted remaining time preempts the currently running request. In contrast, Trail disables preemption once a request reaches a certain ``age'' (i.e., a fraction of its predicted total work). The intuition is that during the early, or ``young'' phase of execution, the KV cache is small, making preemption relatively inexpensive in terms of memory overhead. Later, in the ``old'' phase, a significant amount of memory has been allocated for the KV cache, so it becomes more efficient to complete the request rather than preempt it and later reallocate the required memory. Consequently, Trail enforces a global threshold of \(c \cdot \text{predicted request size}\) (with \(0 \leq c \leq 1\)), disabling preemption once a request’s age exceeds this threshold. 
Trail reduces mean latency by 1.66× to 2.01× on the Alpaca dataset~\cite{alpaca} and achieves 1.76× to 24.07× lower mean time to the first token compared to vLLM, evaluated on a server with a single NVIDIA A100 80GB GPU.
We note that a simplified M/G/1 policy that utilizes the idea of Trail, disabling preemptions at a threshold of \(c \cdot \text{predicted request size}\), can be analyzed using SOAP methods, as shown in \cite{trail}.  Such analyses can give insight into the tradeoffs of SPRPT variants when memory or other issues need to be taken into account.


\medskip

\noindent{\bf Open Questions:}

\begin{itemize}

    \item How should we rank requests given the split between prefill and decode stages? As a challenging example, consider two requests with the same total size but different prefill and decode phase sizes (e.g., one with short prefill and long decode versus one with a long prefill and short decode).  The appropriate ranking may depend on the hardware environment and whether split-phase scheduling is used for the phases.
    
    \item {How can preemption thresholds be dynamically adjusted?}
    Since the effects of preemption depend on factors such as model size, batch size, available memory, and the distribution of incoming requests, what strategies can dynamically tune preemption thresholds based on the sizes (or expected sizes) of batched requests?

    \item  Are there other interesting variants of SPRPT with limited preemptions worth studying, and how do they affect other performance measures (such as tail behavior)?
\end{itemize}

\subsection{Adaptive Scheduling}

While job-level scheduling optimizes the execution of individual requests, many systems operate on usage-based billing models, making operating budgets a crucial design factor. From a system-level perspective, adaptive scheduling in LLM deployments must address cost constraints, heterogeneous hardware, and prompt sharing. Incorporating financial considerations into scheduling algorithms leads to multi-objective optimization formulations that balance low latency with cost efficiency, introducing trade-offs between latency and cost.


Further gains may be achieved by incorporating prompt sharing, which occurs when multiple requests contain overlapping input segments, enabling the system to reuse intermediate KV computations and reduce redundant processing during inference.
In many LLM applications, prompts overlap across user requests and often share common prefixes. For example, \cite{srivatsa_preble_2024} reports that 85\% to 97\% of tokens in a prompt may be shared with other prompts. Such sharing occurs in settings such as conversational agents~\cite{anthropic_prompt_2024}, tool use~\cite{qin_toolllm_2023, schick_toolformer_2023}, question answering~\cite{li_loogle_2024,rawal_cinepile_2024, wang_augmenting_2024}, complex reasoning~\cite{besta_graph_2024, wei_chain--thought_2024}, batch inference~\cite{li_competition-level_2022}, in-context learning~\cite{dong_survey_2024}, and agent systems~\cite{chen_reconcile_2024,gao_empowering_2024,guo_large_2024}. \cite{juravsky_hydragen_2024} and \cite{srivatsa_preble_2024} explore methods that leverage shared prompts to improve online LLM inference efficiency. These methods reuse common prefixes to reduce redundant computation. However, these systems focus on reuse of KV cache entries and balance computational load across GPUs using data parallelism without explicitly optimizing for latency. SGLang~\cite{zheng2025sglang} introduces RadixAttention for efficient KV cache reuse across requests. Instead of discarding the KV cache after each request, SGLang stores both prompts and outputs in a radix tree that supports fast prefix search, insertion, and eviction using an LRU policy. They show settings where this approach yields up to 5 times higher throughput than vLLM.

Shared prompts can reduce the cost of the prefill phase when requests sharing the same context are batched; however, it remains unclear how best to order such requests. For instance, consider three requests \(R_1\), \(R_2\), and \(R_3\), where \(R_3\) is small and \(R_1\) and \(R_2\) share a long context. A naive strategy might always prioritize the smaller \(R_3\), but that approach misses the opportunity to batch \(R_1\) and \(R_2\) together. Thus, there is a need to develop adaptive algorithms that continuously weigh prompt overlap, real-time GPU load, and queue dynamics to minimize both latency and resource underutilization. Theoretical modeling and trade-off analysis of these multi-factor schedulers represent an important challenge in large-scale LLM deployments.

Another challenge is addressing the divergent latency requirements of interactive and batch requests. Interactive requests require near-immediate responses, whereas long-running or batch requests can tolerate delays but must avoid starvation. Prioritizing short, interactive queries improves responsiveness but risks indefinitely postponing larger tasks under high load. SAGESERVE~\cite{jaiswal2025serving} presents a system 
for serving LLM inference requests with a wide range of service level agreements (SLAs), which maintains better GPU utilization and reduces resource fragmentation that occurs in isolated resource pools (or silos).
The goal is to develop scheduling algorithms that ensure low latency for interactive queries while maintaining the progress of non-interactive workloads. This challenge can be addressed within a multi-objective scheduling framework that balances latency and fairness, using queueing theory to prevent starvation.

\medskip

\noindent{{\bf Open Questions:}}
\begin{itemize}
    \item How can we develop scheduling and resource-allocation strategies that meet a global cost target while ensuring acceptable response times? In particular, we may seek to 
    design an adaptive approach that dynamically adjusts the prioritization of latency and cost based on workload characteristics and SLA requirements.

    \item {How should scheduling policies handle interactive and non-interactive workloads?} 

    \item {How should prompt sharing be incorporated into scheduling decisions?}
    In scenarios where requests share similar prompts, what strategies can be employed to batch such requests effectively while avoiding delays for standalone small requests?
\end{itemize}

\subsection{GPU Resource Allocation}

\begin{figure}[H]
    \centering
    \begin{subfigure}[b]{0.4\linewidth}  
        \centering
        \includegraphics[width=\linewidth]{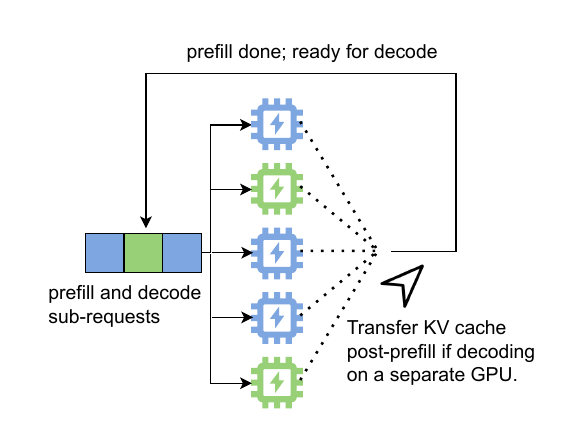}
        \caption{Pooled GPUs}
        \label{fig:pooled_gpus}
    \end{subfigure}
    \begin{subfigure}[b]{0.4\linewidth}  
        \centering
        \includegraphics[width=\linewidth]{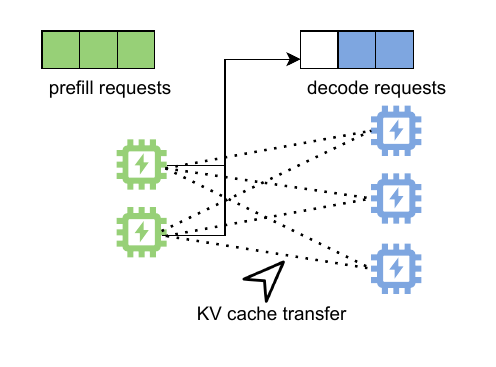}
        \caption{Dedicated GPUs for prefill and decode}
        \label{fig:dedicated_gpus}
    \end{subfigure}
    \vspace{-8pt} 
    \caption{GPU organizations: comparing pooled GPUs with dedicated GPUs for prefill and decode phases.}
    \label{fig:gpus_organization}
\end{figure}

Shifting to the system perspective, we now examine how to orchestrate GPU resources, each with varying computational and memory capabilities, across various architectures. For example, GPUs optimized for \emph{decode} may favor larger memory capacities with relatively lower compute throughput, while those optimized for \emph{prefill} may prioritize higher compute throughput with lower memory footprints. 

Figure~\ref{fig:gpus_organization} shows two organizations. In the \emph{pooled GPUs} approach (Figure~\ref{fig:pooled_gpus}), each GPU handles both the prefill and decode phases. After completing the prefill phase, a request can either proceed directly to decoding on the same GPU or re-enter the queue to be processed by another available GPU. The latter option poses a challenge: transferring the KV cache from the prefill GPU to a decode GPU and storing it until the assignment introduces memory and communication overhead. This raises the question of whether to complete the request immediately or preempt it and rank it among waiting requests. 
The \emph{dedicated GPUs} approach (Figure~\ref{fig:dedicated_gpus}) partitions the pool of GPUs into separate groups: one exclusively handling prefill and the other exclusively handling decode. This arrangement is similar to tandem servers in queueing systems \cite{burke1956output,reich1957waiting}, but here the GPUs can differ in speed and require KV cache transfer, complicating scheduling further. Although this approach can mitigate conflicts within each phase, it risks underutilization of one subset of GPUs if workload distributions are imbalanced (for example, when large prompt sizes prolong prefill while decode resources remain idle). In heterogeneous GPU environments, mismatches between GPU capabilities and different phases (prefill vs. decode) requirements can lead to inefficient resource usage and, thus, higher operational costs. Effective scheduling must balance these phases to optimize both cost and latency.

Dynamic hardware availability further complicates scheduling decisions. In many practical settings, additional GPUs become available after an LLM service is already running. The challenge is to integrate these resources without disrupting ongoing workloads and to determine whether they should support the prefill phase to improve the handling of large prompts, or the decode phase to reduce response times for interactive queries. Bottlenecks, workload composition, and cost constraints may change over time, making static allocations suboptimal.

\medskip

\noindent{\bf Open Questions:}

\begin{itemize}
    \item How should GPU resources be orchestrated across pooled and dedicated organizations?

    \item How can scheduling policies be designed to balance prefill and decode phases when GPUs have heterogeneous compute and memory capabilities? 

    \item Can results or insights from tandem queues give insight into optimizing the dedicated GPUs approach?

    \item Can we systematically estimate a number of lower-capacity GPUs, each with limited memory and throughput, required to match or exceed the performance of a single high-end GPU? 
    We may seek to do this because of significant price differentials between low and high-end GPUs.
    Beyond raw throughput, this evaluation must account for communication overhead between GPUs, the availability and cost of high-bandwidth interconnects, and cumulative power. Developing theoretical models and empirical studies of these factors will clarify when a single, more capable GPU is preferable to a cluster of smaller GPUs, especially for the prefill and decode phases.

    \item How can we dynamically identify resource-constrained phases and design scheduling algorithms that adapt to changes in GPU availability? 
    One natural approach is to use predictions of request arrival rates and prompt sizes to allocate resources to the most critical phase.
    
\end{itemize}

\section{Scheduling in Compound AI Systems}
\label{sec:compound_systems}

In the previous section, we focused on a single LLM deployment. However, AI development is shifting toward compound systems that integrate multiple interacting components -- such as external tools, model calls, and retrievers -- rather than relying on monolithic models. LLMs are typically trained on general pretraining datasets consisting of short text, which do not provide high-quality examples for tasks requiring long contexts or frequent knowledge updates. Retrieval-Augmented Generation (RAG) addresses this limitation by incorporating an information retrieval step to enhance the generation process. In a typical RAG workflow, a retriever identifies relevant data sources for a given request, and the retrieved information is integrated with the input, leading to higher accuracy and better robustness. Surveys~\cite{gao2023retrieval, zhao2024retrieval} include details on RAG.

LLMs can orchestrate external API (Application Programming Interfaces) calls to fetch up-to-date information or perform computations, and agent-based approaches enable LLMs to autonomously plan and execute tasks across various specialized modules. As these compound systems become more prevalent, improving their efficiency and adaptability becomes critical.
This section presents the scheduling problem within such systems.

\subsection{Augmented LLMs}

\begin{figure}[H]
  \centering
  \includegraphics[width=0.5\linewidth]{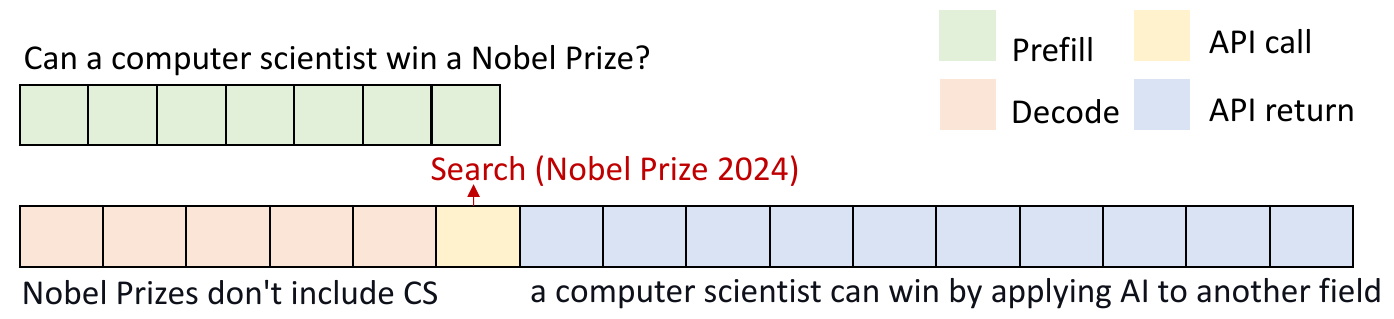}
  \caption{Illustration of an augmented-LLM request. The API fetches detailed information about the 2024 Nobel Prize.}
  \label{fig:augmented_llm}
\end{figure}

Augmented Language Models~\cite{mialon2023augmented, wang2024survey} enhance traditional LLM capabilities by incorporating external tools or retrieval mechanisms. Unlike pure LLMs that rely solely on pre-trained parameters to generate responses, augmented LLMs can query external data sources to expand their functionality. Figure~\ref{fig:augmented_llm} illustrates an example of an augmented LLM request. These augmentations, here referred to as \emph{API}, fall into three main categories as described in~\cite{mialon2023augmented}: incorporating non-LLM tools during decoding (such as calculators~\cite{wolfarm}, information retrieval systems~\cite{baeza1999modern}), iterative self-calling of an LLM (like chatbots maintaining conversation history), and complex compositions involving multiple LLMs, models, and tools (exemplified by frameworks like LangChain~\cite{langchain}, DSPy~\cite{khattab2024dspy}, Gorilla~\cite{patil2023gorilla}, SGLang~\cite{zheng2023efficiently}, and AgentGraph~\cite{chen2019agentgraph}).

API call durations vary significantly between augmentation types, distinguishing short-running from long-running augmentations. Thus, API handling strategies should be tailored to the augmentation type rather than adopting a one-size-fits-all approach. During an API call, there are three primary strategies for handling the request:

\begin{itemize} 

    \item \emph{Preserve}: Retain the KV cache in memory while waiting for the API response.

    \item \emph{Discard and Recompute}: Remove the KV cache and rebuild it once the API returns.

    \item \emph{Swap}: Offload the KV cache to CPU memory and reload it when the API returns.

\end{itemize}

\begin{figure}[H]
    \centering
    \begin{subfigure}[b]{0.25\linewidth}
        \centering
        \includegraphics[width=\textwidth]{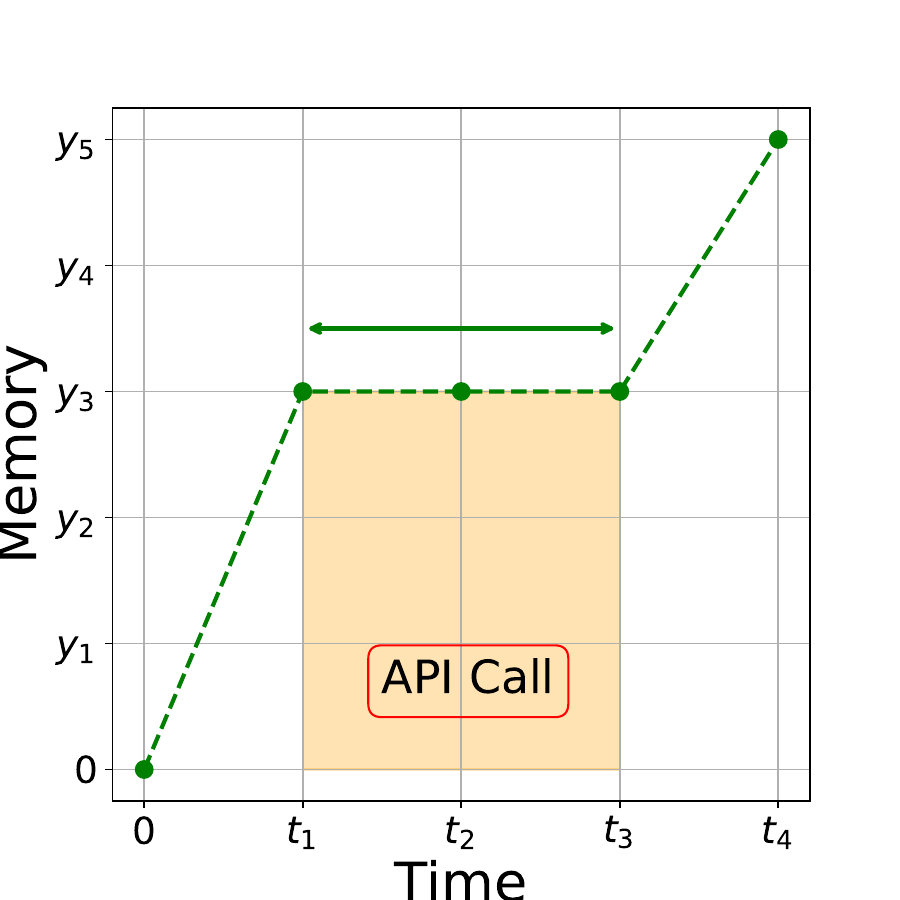}  
        \caption{Preserve}
        \label{fig:preserve_over_time}
    \end{subfigure}
    \hfill
    \begin{subfigure}[b]{0.25\linewidth}
        \centering
        \includegraphics[width=\textwidth]{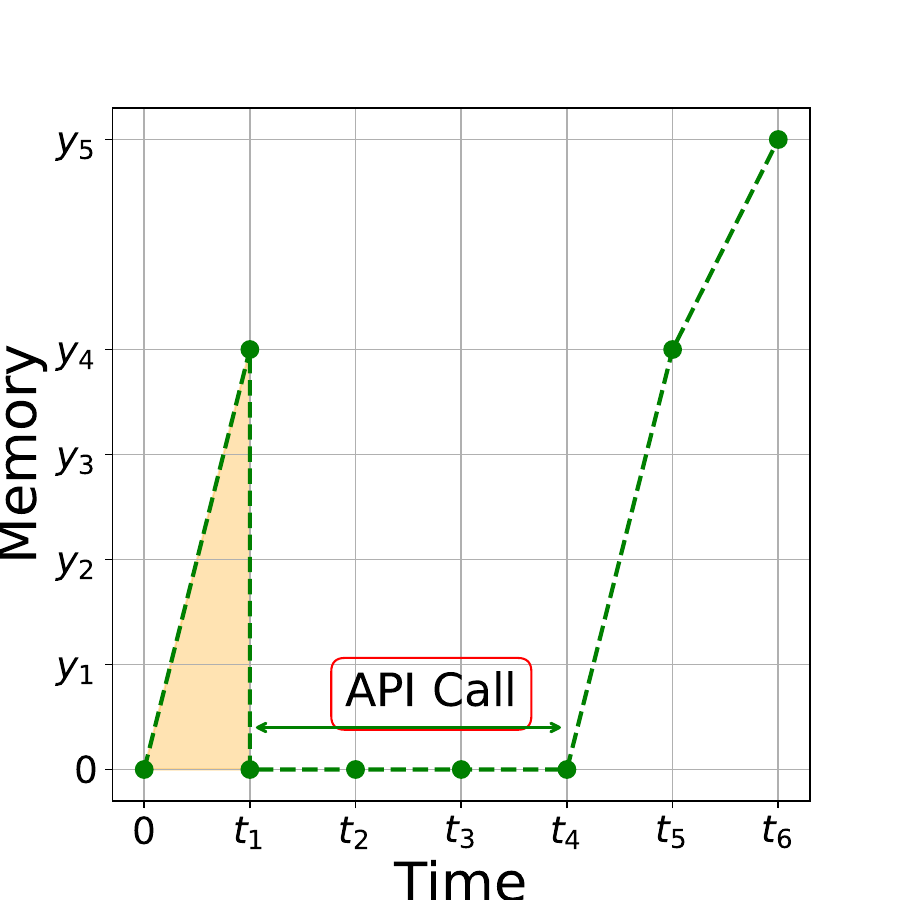}  
        \caption{Discard}
        \label{fig:recompute_over_time}
    \end{subfigure}
    \hfill
    \begin{subfigure}[b]{0.25\linewidth}
        \centering
        \includegraphics[width=\textwidth]{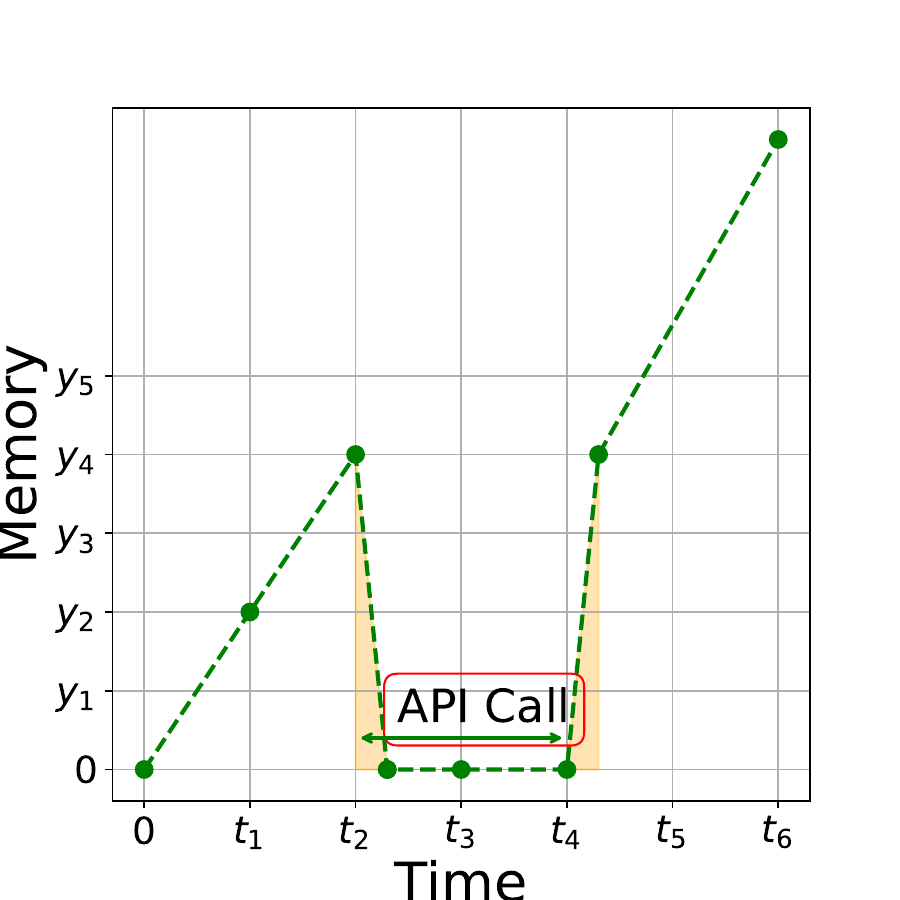}  
        \caption{Swap}
        \label{fig:swap_over_time}
    \end{subfigure}
    \caption{Memory consumption over time for a request with one API call using three memory management strategies: (1) Preserve, (2) Discard and Recompute, and (3) Swap. The highlighted area represents memory waste for one request.}
    \label{fig:memory_over_time}
\end{figure}

Each strategy has drawbacks. With \emph{Preserve}, the KV cache occupies memory even while the LLM is idle. \emph{Discard and Recompute} wastes both memory and compute resources during cache reconstruction. \emph{Swap} incurs data-transfer overhead and pauses request processing while transferring the KV cache. Figure~\ref{fig:memory_over_time} illustrates the memory-over-time function, with the highlighted areas in Figures~\ref{fig:preserve_over_time},~\ref{fig:recompute_over_time}, and~\ref{fig:swap_over_time} indicating the memory waste for a single request due to an API call.

API calls can range from milliseconds for simple calculations to several seconds for complex tasks like image generation. API-augmented requests challenge the system by increasing KV cache memory demands during the memory-bound LLM decoding phase. INFERCEPT~\cite{abhyankarinfercept} proposes a method to determine the handling strategy when a request reaches an API call; however, it lacks integrated scheduling policies to proactively minimize latency and relies on a FIFO approach, which may lead to head-of-line blocking.

Size-based scheduling methods can typically reduce request response times by utilizing known or predicted request sizes. Traditional scheduling, however, faces challenges with API-augmented requests because their memory requirements do not scale proportionally with execution time. In this context, it is unclear whether the API delay should be included in the size estimate. Even with known output sizes, techniques such as shortest job first may fail to perform effectively when requests involve API calls.

The work in~\cite{fast_augmentedLLMs} is the first to propose scheduling policies beyond FIFO for augmented LLMs, presenting a system called LAMPS. LAMPS employs a predictive, memory-aware scheduling approach that integrates API handling with request prioritization to reduce completion times. It operates in two steps: First, it assigns a handling strategy to API-augmented requests before scheduling, based on predictions for expected output size and API call duration. Then, it schedules requests by ranking them according to their predicted total memory footprint across their lifecycle, factoring in both request size and API interactions. LAMPS achieves end-to-end latency improvements of 27\%-85\% and reductions in time-to-first-token (TTFT) of 4\%-96\% compared to INFERCEPT, with even greater gains over vLLM, an LLM serving system that applies paged attention to reduce memory overhead.

\medskip

\noindent{\bf Open Questions:}

\begin{itemize}
    \item {What theoretical guarantees can scheduling algorithms offer for API-augmented requests?}
    Few theoretical models address scheduling for the types of API calls examined here. Investigating algorithmic bounds, such as competitive ratios or approximation guarantees, for even simplified models may reveal new insights and guide the development of practical scheduling solutions.


\item How can load balancing be implemented to route LLM inference requests among machines hosting API endpoints while considering real-time load with possible quality tradeoffs?
\end{itemize}

\subsection{Multiple LLMs Available}

\begin{figure}[H]
\centering
\includegraphics[width=0.6\linewidth]{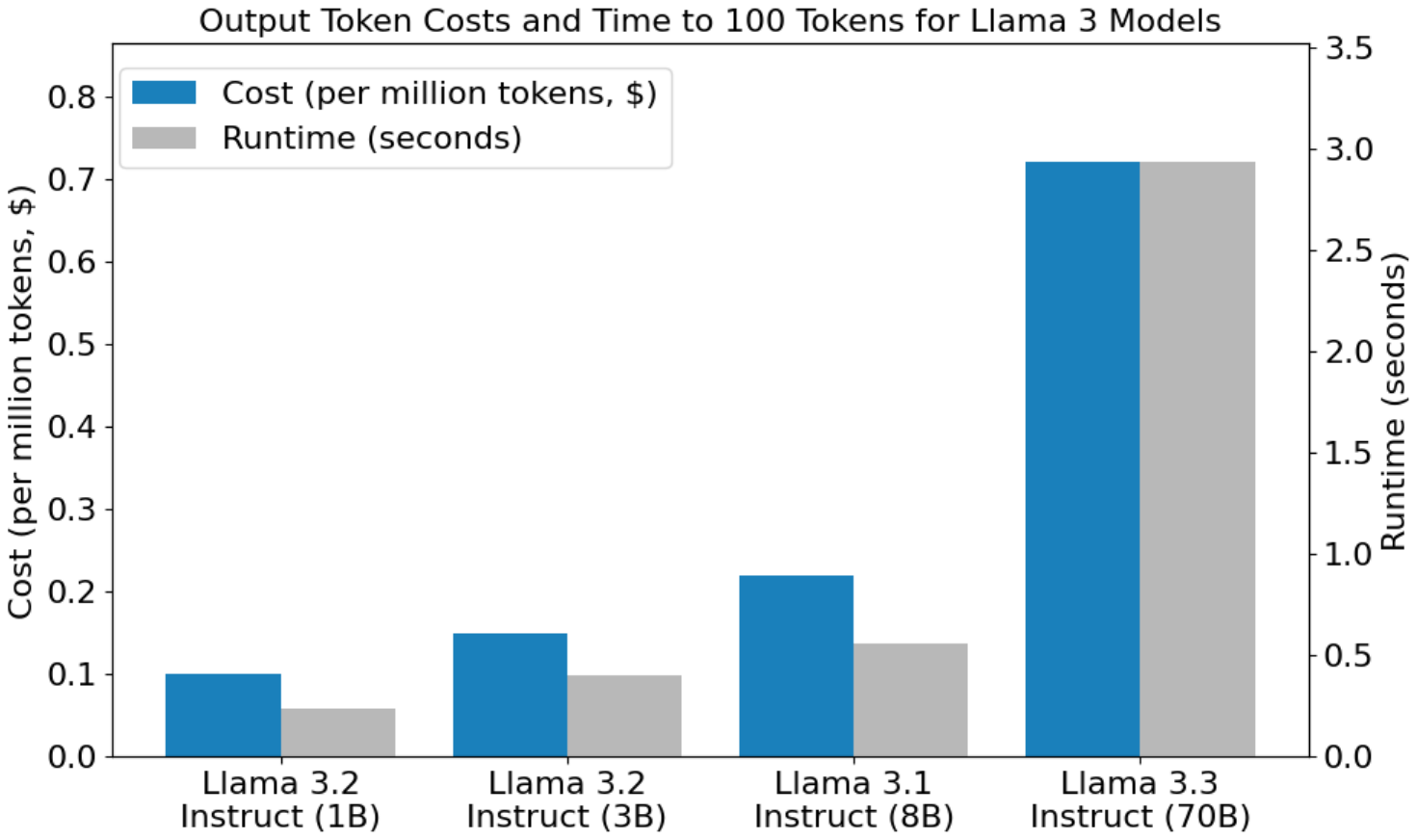} \caption{Comparison of output token costs (per million tokens) for the Llama 3 family on Amazon Bedrock and the time (in seconds) required to generate 100 tokens, measured on two NVIDIA H100 GPUs.}
\label{fig:llama-cost-latency}
\end{figure}

AI development is moving toward compound AI systems~\cite{zaharia2024} that integrate multiple components, often including LLMs of varying sizes and capabilities. Model size significantly impacts runtime, cost, and answer quality in transformer-based LLMs, as shown in Figure~\ref{fig:llama-cost-latency}. Smaller models may suffice for straightforward queries, while larger models provide deeper answers to complex prompts.

Speculative decoding~\cite{chen2023accelerating, leviathan2023fast}, also known as speculative sampling, employs two LLMs of different sizes (small and large) simultaneously to accelerate token generation while maintaining accuracy. This approach can reduce inference time by 2.5 to 3 times.
The key insight is that the transformer architecture allows a single forward pass to generate one token or evaluate multiple tokens in parallel. Consequently, the computational cost of generating one token is equivalent to that of verifying an entire sequence of tokens concurrently.
In this method, the smaller LLM generates preliminary tokens that the larger LLM verifies to determine which tokens to accept. When a token is rejected based on a comparison of the probability distributions from both models, the next token is sampled from the larger model. This process preserves overall accuracy (compared to using only the large model) through the speculative approach.
An interpretation of this process is that the tokens generated by the smaller model are preliminary predictions that the larger model subsequently verifies and, if necessary, corrects.
This can be viewed through the analogy of a tandem model, where performance depends on the quality of the initial prediction. The time spent in the ``first queue'' (the small model) directly impacts the workload of the ``second queue'' (the large model). In other words, the accuracy of the small model’s predictions determines the additional processing required by the larger model.

Routing all requests to a single large model can incur high costs, load, and latency, which degrades user experience and limits real-time applications. For example, on two NVIDIA H100 GPUs, Llama 3.2 Instruct (1B) generates 100 tokens in 0.2 seconds, whereas Llama 3.3 Instruct (70B) requires 3 seconds, reflecting greater computational complexity. Similarly, cost differences are significant; Amazon Bedrock charges \$0.10 per million tokens for Llama 3.2 Instruct (1B) compared to \$0.72 for Llama 3.3 Instruct (70B).\footnote{\href{https://aws.amazon.com/bedrock/pricing/}{Amazon Bedrock pricing} is for the \texttt{us-east-1} region, as of February 2025.} These observations motivate the need for advanced load balancing and scheduling strategies in compound AI systems.

A key challenge is identifying the appropriate LLM to query based on the desired answer quality. Previous work~\cite{ong2024} proposes four methods for predicting response quality in a two-model setting: similarity-weighted matching to training set labels, matrix factorization, a BERT-based classifier, or a call to Llama 3.1 Instruct (8B). These predictors are trained on human preference data augmented with LLM judge-labeled datasets.

\medskip

\noindent{\bf Open Questions:}
    \begin{itemize}
    \item Can we model the speculative decoding process as a tandem queue, and use it to gain insights into how to optimize the use of the smaller model?
    \item How can scheduling algorithms dynamically assign incoming requests to an appropriate LLM based on predicted response size and quality, while meeting user-specified cost and latency constraints?
    
    \item 
    Given that observed performance may differ from predictions under unpredictable workloads,
    how can scheduling algorithms adaptively balance predictions with observed performance (real-time feedback) under unpredictable workloads?
     \item Can queueing models give insight into optimizing the tradeoffs between accuracy of results and the time spent in the system for LLM systems?
\end{itemize}

\subsection{Multiple LLMs Needed}
In the previous section, we discussed scenarios involving a single LLM selected from a pool of different LLM sizes to serve a request. Many compound AI systems, however, require sequential processing by multiple LLMs. We can view this setting from two primary perspectives. In the first, each LLM is responsible for a specific subtask, with a central coordinator aggregating the agents' outputs to form a final response. In the second, the processing follows a directed acyclic graph (DAG) of sub-jobs, where each stage processes and refines the previous stage's output, iteratively developing the response.

Both approaches present significant research challenges. While aiming to optimize latency and reduce overhead, they differ fundamentally in task dependencies and scheduling complexities. In the agent-based model, LLMs operate as independent agents processing assigned subtasks. The scheduling challenge here centers on allocating incoming requests across heterogeneous GPU resources to maximize parallelism and throughput while minimizing coordination overhead during output aggregation.
Alternatively, the DAG-based approach structures processing as a sequence of interdependent stages, with each stage's output serving as the subsequent stage's input. This model demands scheduling strategies that carefully manage stage dependencies, balancing pipeline parallelism with the synchronization needed for smooth inter-stage transitions. The objective is to minimize end-to-end latency despite variability in execution times and communication overhead.
Though both settings share the overarching goals of reducing latency and overhead, they differ in fundamental approaches: the agent-based model emphasizes the parallel distribution of independent tasks, while the DAG-based model concentrates on orchestrating a chain of dependent operations.

\medskip

\noindent{\bf Open Questions:}
\begin{itemize}
    \item How can we design scheduling algorithms for multi-LLM systems, both in the setting of independent, parallel, and in the setting of processing sequential, interdependent tasks?
    How can predictions aid scheduling decisions in these types of systems
    
    
    \item What theoretical models from queueing theory and job-shop scheduling can be extended to provide either performance guarantees for (possibly simplified models of) multi-stage LLM inference systems or insights for scheduling approaches for such systems.
\end{itemize}





\section{Scheduling in LLM Reasoning Systems}
\label{sec:llm_reasoning}



LLMs can now perform advanced reasoning tasks such as solving mathematical problems, generating code, and analyzing legal documents. Inference-time reasoning algorithms play a key role by allowing LLMs to evaluate their outputs, explore alternative reasoning paths, and produce more reliable responses to complex questions.
LLM reasoning algorithms have two fundamental phases: \emph{expansion}, in which the model generates tokens to explore different solution paths, where allocating more compute to expansion improves answer quality, and \emph{aggregation}, where these candidate solutions are combined to produce a final answer.

LLM reasoning can be approached using the following distinct approaches.
One popular approach is the majority~\cite{wang2022self} (or self-consistency), where a prompt is processed multiple times with different random seeds or temperature settings. The final output is determined by majority voting across these candidate responses, which enhances robustness by mitigating errors from any single inference run. The rebase~\cite{wu2024inference} approach generates multiple intermediate reasoning steps from a given prompt. A reward model scores these steps, and the highest-scoring nodes are selected to guide further expansion. This iterative process constructs a reasoning tree, resulting in a final answer obtained either through weighted majority voting or by selecting the top-scored candidate. Another approach is Monte Carlo tree search~\cite{feng2023alphazero, hao2023reasoning}, which builds a solution tree by iteratively expanding nodes. At each step, a continuation is sampled from the LLM until a leaf node with a candidate solution is reached; the candidate’s score is then back-propagated to update its ancestors, enabling effective exploration of both depth and breadth in the solution space. Finally, the internalized chain-of-thought leverages modern LLMs~\cite{openai_o1, qwq, guo2025deepseek} that have been trained to generate extended chain-of-thought~\cite{wei2022chain} sequences in a single pass. These models internally refine their outputs by incorporating intermediate reasoning steps until reaching a final answer, eliminating the need for explicit multi-run aggregation or tree search.

Prior works on serving systems~\cite{agrawal2024taming, patel2024splitwise, kwon2023efficient, trail, fu2024efficient} assume independent input/output requests and have extensively optimized LLM inference at the system level. These systems, however, overlook LLM reasoning programs that may submit interdependent inference requests. Parrot~\cite{lin2024parrot} and SGLang~\cite{zheng2025sglang} both target multi-request applications. Parrot offers an abstraction that allows users to specify dependencies between requests, enabling more effective cross-request scheduling. SGLang introduces programming primitives tailored for multi-request workflows, optimizing execution by reusing intermediate KV cache memory across requests. However, these systems do not specifically address LLM reasoning programs. A recent work, Dynasor~\cite{fu2024efficiently} targets reasoning systems by optimizing inference-time computing for LLM reasoning queries. It allocates additional compute resources to challenging queries, reduces compute for simpler ones, and terminates unpromising queries early. This balances accuracy, latency, and cost.

Reasoning systems demand adaptive scheduling that responds dynamically to the evolving state of the reasoning process. While some queries converge after only a few steps, others require extensive exploration, leading to execution times that conventional scheduling algorithms are not designed to handle. In this context, the notion of request size goes beyond the number of output tokens to include the number of reasoning steps needed for convergence. This additional information can be predicted and used to inform more effective scheduling decisions.

LLM reasoning systems thus present a rich ground for developing adaptive scheduling algorithms that allocate resources based on intermediate reasoning outputs. For example, if early reasoning steps indicate a clear and promising trajectory toward a correct answer, the scheduler might continue to invest in additional computational resources. Conversely, if early signals suggest divergence, the system could preempt the process and reallocate resources to other tasks. Moreover, there is potential for exploring parallel evaluation of multiple reasoning paths, which raises further challenges regarding load balancing, redundancy, and optimal resource sharing.


\medskip

\noindent{\bf Open Questions:}
\begin{itemize}
\item Since current reasoning algorithms share the core phases of expansion and aggregation, can we design a unified scheduling framework that dynamically allocates resources based on the current phase? In particular, how should scheduling priorities and resource management differ between the expansion phase and the aggregation phase?
\item How can we develop a reasoning system that leverages KV cache sharing among different reasoning paths within a single request?
\item How can pre-run predictions of reasoning complexity (depth) be used to inform scheduling decisions, and how should we mitigate prediction errors?
\end{itemize}
\section{Conclusion}
\label{sec:conclusion}
Advances in algorithms with predictions have demonstrated the benefits of integrating machine learning with classical algorithms across various domains. Queueing systems are one such area, where recent works explore how predictions of service times can optimize scheduling. However, key questions remain regarding the limitations of existing theoretical frameworks and the robustness of queueing models under different predictive assumptions.

Interestingly, scheduling algorithms using predictions have already been shown to have significant potential for improving performance of LLM systems, particularly for inference. However, LLMs introduce particular scheduling challenges due to their memory-intensive inference processes, the need for dynamic batching, and the impact of preemption on KV cache management. Unlike traditional queueing systems, LLM systems must also account for factors such as cost and answer quality, and they involve multiple processing phases with distinct resource requirements that standard models do not capture. The growing complexity of AI deployments has further led to the emergence of LLM systems that extend beyond single-instance setups. These include compound AI platforms that integrate multiple LLMs with external tools, as well as reasoning systems. Each of these architectures introduces additional challenges that provide new questions for queueing theory to consider.  

Our goal in this paper is to highlight key challenges and open questions in algorithms with prediction in queueing in general, and to describe particular new problems that arise in the context of LLMs that could potentially benefit from theoretical and algorithmic insights. Specifically, we highlight the need for new theoretical models that accommodate the characteristics of LLM inference. Addressing these challenges requires rethinking scheduling algorithms to better adapt to the growing complexity of modern AI systems.

\section*{Acknowledgements}
Michael Mitzenmacher was supported in part by NSF grants CCF-2101140, CNS-2107078, and DMS-2023528.
Rana Shahout was supported in part by the Schmidt Futures Initiative, the Zuckerman Institute, and NSF grants CNS-2107078 and DMS-2023528.

\pagebreak

\ifdefined \arvix
\bibliographystyle{plain}
\else
\bibliographystyle{ACM-Reference-Format}
\fi

\bibliography{refs_new}

\appendix


\end{document}